\documentclass{article}

\usepackage{ulem}
\usepackage{hyperref}
\usepackage{url}
\usepackage{multirow}
\usepackage{rotating}
\usepackage{wrapfig}
\usepackage{xspace}
\usepackage{wrapfig}
\usepackage{subfig}
\usepackage{amsmath}

\usepackage{hyperref}

\newcommand{\framework}{InfoTime}
\newcommand{\mf}{CDAM}
\newcommand{\ms}{TAM}

\usepackage[accepted]{icml2024}

\usepackage{amsmath}
\usepackage{amssymb}
\usepackage{mathtools}
\usepackage{amsthm}
\usepackage{amsmath}

\usepackage[capitalize,noabbrev]{cleveref}

\theoremstyle{plain}

\theoremstyle{definition}

\theoremstyle{remark}

\newcommand{\E}{\mathbb{E}}

\usepackage[textsize=tiny]{todonotes}

\icmltitlerunning{Submission and Formatting Instructions for ICML 2024}

\begin{document}

\twocolumn[
\icmltitle{Enhancing Multivariate Time Series Forecasting with Mutual Information-driven Cross-Variable and Temporal Modeling}






\icmlsetsymbol{equal}{*}

\begin{icmlauthorlist}
\icmlauthor{Shiyi Qi}{loc:a}
\icmlauthor{Liangjian Wen}{loc:b}
\icmlauthor{Yiduo Li}{loc:a}
\icmlauthor{Yuanhang Yang}{loc:a}
\icmlauthor{Zhe Li}{loc:a}
\icmlauthor{Zhongwen Rao}{loc:b}
\icmlauthor{Lujia Pan}{loc:b}
\icmlauthor{Zenglin Xu}{loc:a}
\end{icmlauthorlist}

\icmlaffiliation{loc:a}{Harbin Institute of Technology, Shenzhen}
\icmlaffiliation{loc:b}{Huawei Technologies Ltd.}

\icmlcorrespondingauthor{Shiyi Qi}{21s051040@stu.hit.edu.cn}
\icmlcorrespondingauthor{Liangjian Wen}{wenliangjian1@huawei.com}
\icmlcorrespondingauthor{Yiduo Li}{lisa\_liyiduo@163.com}
\icmlcorrespondingauthor{Yuanhang Yang}{ysngkil@gmail.com}
\icmlcorrespondingauthor{Zhe Li}{plum271828@gmail.com}
\icmlcorrespondingauthor{Zhongwen Rao}{raozhongwen@huawei.com}
\icmlcorrespondingauthor{Lujia Pan}{panlujia@huawei.com}
\icmlcorrespondingauthor{Zenglin Xu}{zenglin@gmail.com}

\icmlkeywords{Machine Learning, ICML}

\vskip 0.3in
]



\printAffiliationsAndNotice{}  

\begin{abstract}
Recent advancements have underscored the impact of deep learning techniques on multivariate time series forecasting (MTSF).
Generally, these techniques are bifurcated into two categories: Channel-independence and Channel-mixing approaches. Although Channel-independence methods typically yield better results, Channel-mixing could theoretically offer improvements by leveraging inter-variable correlations. Nonetheless, we argue that the  integration of uncorrelated information in channel-mixing methods could curtail the potential enhancement in MTSF model performance. To substantiate this claim, we introduce the \textbf{C}ross-variable \textbf{D}ecorrelation \textbf{A}ware feature \textbf{M}odeling (\mf) for Channel-mixing approaches, aiming to refine Channel-mixing by minimizing redundant information between channels while enhancing relevant mutual information. Furthermore, we introduce the \textbf{T}emporal correlation \textbf{A}ware \textbf{M}odeling (\ms) to exploit temporal correlations, a step beyond conventional single-step forecasting methods. This strategy maximizes the mutual information between adjacent sub-sequences of both the forecasted and target series. Combining \mf~ and \ms, our novel framework significantly surpasses existing models, including those previously considered state-of-the-art, in comprehensive tests. 
\end{abstract}

\section{Introduction}
Multivariate time series forecasting (MTSF) plays a pivotal role in diverse applications ranging from  traffic flow estimation~\citep{bai2020adaptive}, weather prediction~\citep{chen2021autoformer}, energy consumption~\citep{zhou2021informer} and healthcare~\citep{bahadori2019temporal}. Deep learning has ushered in a new era for MTSF, with methodologies rooted in RNN-based~\citep{franceschi2019unsupervised, ijcai18Liu, salinas2020deepar, rangapuram2018deep} and CNN-based models~\citep{lea2017temporal, lai2018modeling}, that surpass the performance metrics set by traditional techniques~\citep{box2015time}. A notable breakthrough has been the advent of Transformer-based models~\citep{li2019enhancing, zhou2021informer, chen2021autoformer, zhou2022fedformer}. Equipped with attention mechanisms, these models adeptly seize long-range temporal dependencies, establishing a new benchmark for forecasting efficacy. While their primary intent is to harness multivariate correlations, recent research indicates a potential shortcoming: these models might not sufficiently discern cross-variable dependencies~\citep{murphy2022univariate,nie2022time,zeng2022transformers}. This has spurred initiatives to tease out single variable information for more nuanced forecasting.

\begin{figure*}
\centering
\begin{minipage}{1.3\columnwidth}
    \subfloat[\scriptsize The Framework of Channel-Independence and Channel-mixing models]{
      \label{c} 
      \includegraphics[width=1.00\columnwidth]{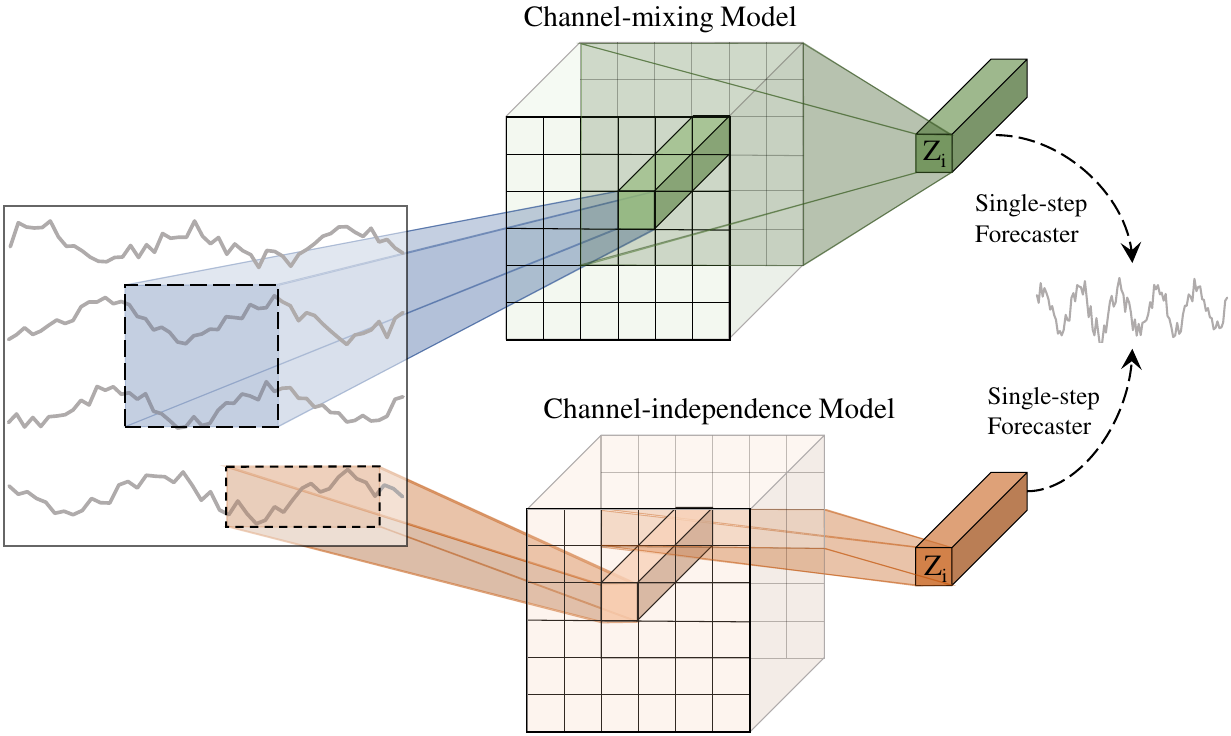}}
\end{minipage}
\begin{minipage}{0.7\columnwidth}
    \subfloat[\scriptsize \centering Traffic flow of 5 adjacent detectors in PEMS08]{
      \label{a} 
      \includegraphics[width=1.0\columnwidth]{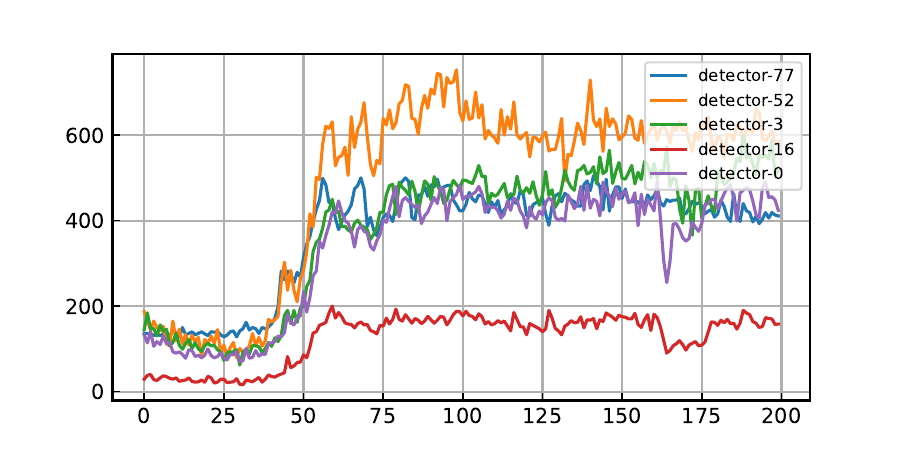}}\\
    \subfloat[\scriptsize \centering Prediction results of PatchTST and Informer in PEMS08]{
      \label{b} 
      \includegraphics[width=1.0\columnwidth]{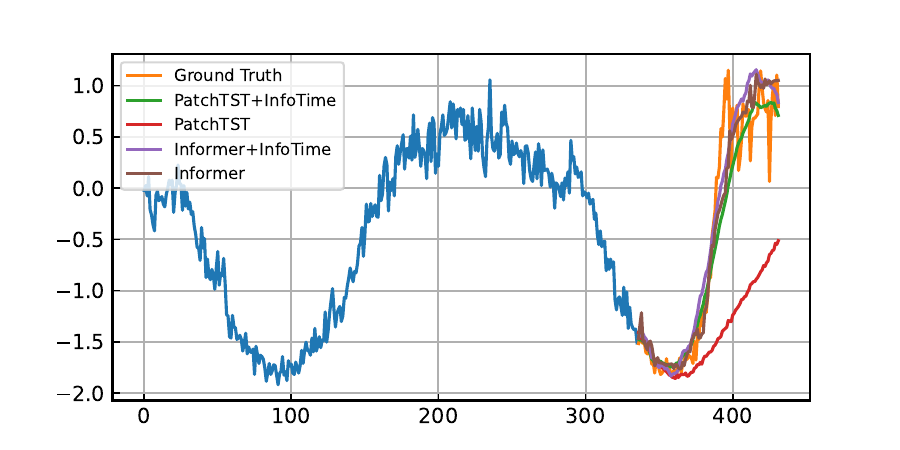}}
\end{minipage}
\caption{(a) The framework of Channel-independence models and Channel-mixing models. Given historical series $X=\{X^i\}$ where $i$ denotes the channel index, the Channel-mixing model tends to maximize the mutual information between $X$ and the latent representation $Z^{i}$. Additionally, it maximize the mutual information between $Z^{i}$ and the i-th future series $Y^{i}$. The Channel-independence models maximize the mutual information between the i-th historical series $X^{i}$ and $Z^{i}$  while ignoring the mutual information between $Z^{i}$ and other channels; (b) Traffic flow of 5 adjacent detectors in the PEMS08 dataset; and (c) Prediction results of 
 Channel-independence model (PatchTST), Channel-mixing model (Informer), and that with our framework, respectively.}
\label{fig:intro_example}
\end{figure*}

When it comes to modeling variable dependencies, MTSF models can be broadly classified into two categories: Channel-mixing models and Channel-independence models, as highlighted in  Figure~\ref{fig:intro_example}~(a)~\citep{nie2022time}. Specifically, Channel-mixing models ingest all features from the time series, projecting them into an embedding space to blend information. Conversely, Channel-independence models restrict their input token to information sourced from just one channel.
Recent studies~\citep{murphy2022univariate,nie2022time,zeng2022transformers} indicates that Channel-independence models significantly outpace Channel-mixing models on certain datasets. Yet, this advantage comes with a trade-off: the omission of crucial cross-variable information. Such an omission can be detrimental, especially when the variables inherently correlate. Illustratively, Figure~\ref{fig:intro_example}~(b) showcases traffic flow variations from six proximate detectors in the PEMS08 dataset~\citep{chen2001freeway} . A discernible trend emerges across these detectors, suggesting that exploiting their interrelated patterns could bolster predictive accuracy for future traffic flows. In a comparative experiment, we trained both a Channel-independence model (PatchTST) and a Channel-mixing model (Informer) using the PEMS08 dataset. The outcome, as visualized in Figure~\ref{fig:intro_example} (c), unequivocally shows Informer's superior performance over PatchTST, underscoring the importance of cross-variable insights. Motivated by these findings, we introduce the \textbf{C}ross-Variable \textbf{D}ecorrelation \textbf{A}ware Feature \textbf{M}odeling (\mf) for Channel-mixing methodologies. \mf~ aims to hone in on cross-variable information and prune redundant data. It achieves this by minimizing mutual information between the latent depiction of an individual univariate time series and related multivariate inputs, while concurrently amplifying the shared mutual information between the latent model, its univariate input, and the subsequent univariate forecast.


Apart from modeling channel dependence, another significant challenge in MTSF is the accumulation of errors along time, as shown in Figure~\ref{fig:intro_example}~(a).  To mitigate this, a number of studies~\citep{nie2022time,zeng2022transformers,zhou2021informer,zhang2023crossformer} have adopted a direct forecasting strategy using a single-step forecaster {that generates multi-step predictions in a single step}, typically configured as a fully-connected network. Although often superior to auto-regressive forecasters, this method tends to neglect the temporal correlations across varied timesteps in the target series, curtailing its potential to capture series inter-dependencies effectively. Drawing inspiration from the notable temporal relationships observed in adjacent sub-sequences post-downsampling~\citep{liu2022scinet}, we propose \textbf{T}emporal Correlation \textbf{A}ware \textbf{M}odeling (\ms), which  iteratively down-samples and optimizes mutual information between consecutive sub-sequences of both the forecasted and target series.

In essence, this paper delves into two pivotal challenges in multivariate time series forecasting: \textbf{cross-variable relationships} and \textbf{temporal relationships}. Drawing inspiration from these challenges, we develop a novel framework, denoted as  \framework \xspace. This framework seamlessly integrates \mf \xspace and \ms. Our paper's key contributions encompass:


\begin{itemize}
    \item We introduce \textbf{C}ross-Variable \textbf{D}ecorrelation \textbf{A}ware Feature \textbf{M}odeling (\mf) designed specifically for Channel-mixing methods. It adeptly distills cross-variable information, simultaneously filtering out superfluous information.
    \item Our proposed \textbf{T}emporal Correlation \textbf{A}ware \textbf{M}odeling (\ms) is tailored to effectively capture the temporal correlations across varied timesteps in the target series.
    \item Synthesizing  \mf \xspace and \ms, we unveil a cutting-edge framework for MTSF, denominated as  \framework.
\end{itemize}

Through rigorous experimentation on diverse real-world datasets, it's evident that our \framework  \xspace consistently eclipses existing Channel-mixing benchmarks, achieving superior accuracy and notably mitigating overfitting. Furthermore, \framework \xspace enhances the efficacy of Channel-Independent models, especially in instances with ambiguous cross-variable traits. 

\section{Related Work}

\subsection{Multivariate Time Series Forecasting}

Multivariate time series forecasting is the task of predicting future values of variables, given historical observations. With the development of deep learning, various neural models have been proposed and demonstrated promising performance in this task. RNN-based~\citep{franceschi2019unsupervised, salinas2020deepar, rangapuram2018deep} and CNN-based~\citep{lea2017temporal, lai2018modeling} models are proposed for models time series data using RNN or CNN respectively, but these models have difficulty in modeling long-term dependency. In recent years, a large body of works try to apply Transformer models to forecast long-term multivariate series and have shown great potential~\citep{li2019enhancing, zhou2021informer, chen2021autoformer, zhou2022fedformer, nie2022time}, Especially, LogTrans~\citep{li2019enhancing} proposes the LogSparse attention in order to reduce the complexity from $O(L^{2})$ to $O(L(\log L)^{2})$. Informer~\citep{zhou2021informer} utilizes the sparsity of attention score through KL-divergence estimation and proposes ProbSparse self-attention mechanism which achieves $O(L\log L)$ complexity. Autoformer~\citep{chen2021autoformer} introduces a decomposition architecture with the Auto-Correlation mechanism to capture the seasonal and trend features of historical series which also achieves $O(L\log L)$ complexity and has a better performance. Afterword, FEDformer~\citep{zhou2022fedformer} employs the mixture-of-expert to enhance the seasonal-trend decomposition and achieves $O(L)$ complexity. The above methods focus on modeling temporal dependency yet omit the correlation of different variables. Crossformer~\citep{zhang2023crossformer} introduces Two-Stage Attention to effectively capture the cross-time and cross-dimension dependency. But the effectiveness of Crossformer is limited. 
Recently, several works~\citep{murphy2022univariate, nie2022time,zeng2022transformers} observe that modeling cross-dimension dependency makes neural models suffer from overfitting in most benchmarks, therefore, they propose Channel-Independence methods to avoid this issue. However, the improvement is based on the sacrifice of cross-variable information. Besides, existing models primarily focus on extracting correlations of historical series while disregarding the correlations of target series.

\subsection{Mutual Information and Information Bottleneck}
Mutual Information (MI) is an entropy-based measure that quantifies the dependence between random variables which has the form:
\begin{equation}
    I(X;Y)=\int p(x,y) log \frac{p(x,y)}{p(x)p(y)}dxdy 
\end{equation}
Mutual Information was used in a wide range of domains and tasks, including feature selection~\citep{kwak2002input}, causality~\citep{butte1999mutual}, and Information Bottleneck~\citep{tishby2000information}. Information Bottleneck (IB) was first proposed by~\cite{tishby2000information} which is an information theoretic framework for extracting the most relevant information in the relationship of the input with respect to the output, which can be formulated as $max \,\, I(Y;Z)- \beta I(X;Z)$. Several works~\citep{tishby2015deep, shwartz2017opening} try to use the Information Bottleneck framework to analyze the Deep Neural Networks by quantifying Mutual Information between the network layers and deriving an information theoretic limit on DNN efficiency. Variational Information Bottleneck (VIB) was also proposed~\citep{alemi2016deep} to bridge the gap between Information Bottleneck and deep learning. In recent years, many lower-bound estimations~\citep{belghazi2018mutual, oord2018representation} and upper-bound estimations~\citep{poole2019variational, cheng2020club} have been proposed to estimate MI effectively which are useful to estimate VIB. Nowadays, MI and VIB have been widely used in computer vision~\citep{schulz2020restricting, luo2019significance}, natural language processing~\citep{mahabadi2021variational, west2019bottlesum, voita2019bottom}, reinforcement learning~\citep{goyal2019infobot, igl2019generalization}, and representation learning~\citep{federici2020learning, hjelm2018learning}. However, Mutual Information and Information Bottleneck are less researched in long-term multivariate time-series forecasting.
\section{Method}
In multivariate time series forecasting, one aims to predict the future value of time series $y_{t}=s_{t+T+1:t+T+P} \in \mathbb{R}^{P \times C}$ given the history $x_{t}=s_{t:t+T} \in \mathbb{R}^{T \times C}$, where $T$ and $P$ refer the number of time steps in the past and future. $C \geq 1$ refers the number of variables. Given time series $s$, we divide it into history set $X=\{ x_{1},...,x_{N} \}$ and future set $Y=\{ y_{1},...,y_{N} \}$, where $N$ is the number of samples. As depicted in Figure~\ref{fig:intro_example} (a), deep learning methods first extract latent representation $Z^{i}$ from $X$ (Channel-mixing) , or $X^{i}$ (Channel-independence) , and subsequently generate target series $Y^{i}$ from $Z^{i}$.
A natural assumption is that these $C$ series are associated which helps to improve the forecasting accuracy. To leverage the cross-variable dependencies while eliminating superfluous information, in Section~\ref{approach:CAM}, we propose the \textbf{C}ross-Variable Decorrelation \textbf{A}ware Feature \textbf{M}odeling (\mf) to extract cross-variable dependencies. Additionally, in section~\ref{approach:TAM}, we introduce \textbf{T}emporal \textbf{A}ware \textbf{M}odeling (\ms) that explicitly models the correlation of predicted series and allows to generate more accurate predictions compared to the single-step forecaster.

\subsection{Cross-Variable Decorrelation Aware Feature Modeling}
\label{approach:CAM}
Recent studies~\citep {nie2022time,zeng2022transformers,zhou2021informer,zhang2023crossformer} have demonstrated that  Channel-independence is more effective in achieving high-level performance than Channel-mixing. 
However, it is important to note that multivariate time series inherently contain correlations among variables. Channel-mixing aims to leverage these cross-variable dependencies for predicting future series, but it often fails to improve the performance of MTSF. One possible reason for this is that Channel-mixing introduces superfluous information.
To verify this, we introduce \mf \xspace to extract cross-variable information while eliminating superfluous information. Specifically, drawing inspiration from information bottlenecks, \mf \xspace maximizes the joint mutual information among the latent representation $Z^{i}$, its univariate input $X^{i}$ and the corresponding univariate target series $Y^{i}$. Simultaneously, it minimizes the mutual information between latent representation $Z^{i}$ of one single univariate time series and other multivariate series input $X^{o}$. This leads us to formulate the following objective:

 \begin{equation}
    \max \,\, I(Y^{i},X^{i};Z^{i})  \,\, s.t. \,\, I(X^{o};Z^{i})\leq { I_{c}},
\end{equation}

where $I(Y^{i},X^{i};Z^{i})$ represents the joint mutual information between the target series $Y^{i}$, the univariate input $X^{i}$, and the latent representation $Z^{i}$. Additionally, $I(X^{o};Z^{i})$ represents the mutual information between the latent representation $Z^{i}$ and the other multivariate input series $X^{o}$. The constraint $I(X^{o};Z^{i})\leq I_{c}$ ensures that the mutual information between $X^{o}$ and $Z^{i}$ is limited to a predefined threshold $I_{c}$.

With the introduction of a Lagrange multiplier $\beta$, we can maximize the objective function for the i-th channel as follows:
\begin{equation}
\label{ib}
    \begin{aligned}
\mathcal{R}_{IB}^{i}&=I(Y^{i},X^{i};Z^{i})- \beta I(X^{o};Z^{i}) \\
        &=I(Y^{i};Z^{i}|X^{i})+ I(X^{i};Z^{i}) -\beta I(X^{o};Z^{i}),
        \end{aligned}
\end{equation}
where $\beta \geq 0$ controls the tradeoff between $I(Y^{i};Z^{i}|X^{i})$, $I(X^{i};Z^{i})$ and $I(X^{o};Z^{i})$, the larger $\beta$ corresponds to lower mutual information between $X^{o}$ and $Z^{i}$,
indicating that $Z^{i}$ needs to retain important information from $X^{o}$ while eliminating irrelevant information to ensure accurate prediction of $Y^{i}$.
However, the computation of mutual information $I(X^{i},Y^{i};Z^{i})$ and $I(X^{o};Z^{i})$ is intractable. Therefore, we provide variational lower bounds and upper bounds for $I(X^{i},Y^{i};Z^{i})$ and $I(X^{o};Z^{i})$, respectively.

\textbf{Lower bound for} $I(X^{i},Y^{i};Z^{i})$. 
The joint mutual information between latent representation $Z^{i}$, i-th historical series $X^{i}$, and i-th target series $Y^{i}$ is defined as (More details are shown in Appendix~\ref{sec:a2.1}):
\begin{equation}
    \begin{aligned}
        I(X^{i},Y^{i};Z^{i})&=I(Z^{i};X^{i}) +I(Z^{i};Y^{i}|X^{i})  \\
        &=\E_{p(z^{i},y^{i},x^{i})} \left[ \log p(y^{i}|x^{i},z^{i})\right] \\
        &+\E_{p(z^{i},x^{i})} \left [ \log p(x^{i}|z^{i}) \right] + H(Y^{i},X^{i}),  \\
    \end{aligned}
\end{equation}
where the joint entropy $H(Y^{i},X^{i})=-\int p(y^{i},x^{i})dx^{i}dy^{i}$ is only related to the dataset and cannot be optimized, so can be ignored. 
Therefore, the MI can be simplified as: 
\begin{equation}
\begin{aligned}
    I(X^{i},Y^{i};Z^{i}) &=\E_{p(z^{i},y^{i},x^{i})} \left[ \log p(y^{i}|x^{i},z^{i})\right] \\
    &+\E_{p(z^{i},x^{i})} \left [ \log p(x^{i}|z^{i}) \right]+\text{constant}.
\end{aligned}
\end{equation}
Since $p(y^{i}|x^{i},z^{i})$ and $p(x^{i}|z^{i})$ are intractable, we introduce $p_{\theta}(y^{i}|z^{i},x^{i})$ and $p_{\theta}(x^{i}|z^{i})$ to be the variational approximation to $p(y^{i}|x^{i},z^{i})$ and $p(x^{i}|z^{i})$, respectively. Thus the variational lower bound can be expressed as (More details are shown in Appendix~\ref{sec:a2.2}):
\begin{align}
    I(X^{i},Y^{i};Z^{i})-\text{constant} &\geq \E_{p(z^{i},y^{i},x^{i})} \left[ \log p_{\theta}(y^{i}|x^{i},z^{i})\right] \nonumber\\
    &+\E_{p(z^{i},x^{i})} \left [ \log p_{\theta}(x^{i}|z^{i}) \right]\nonumber\\
    &= I_v(X^{i},Y^{i};Z^{i}).
\end{align} 
Hence, the maximization of $I(X^{i},Y^{i};Z^{i})$ can be achieved by maximizing $I_v(X^{i},Y^{i};Z^{i})$.
We assume the variational distribution $p_{\theta}(y^{i}|z^{i},x^{i})$ and $p_{\theta}(x^{i}|z^{i})$  as the Gaussion distribution. 
Consequently, the first term of $I_v(X^{i},Y^{i};Z^{i})$ represents the negative log-likelihood of predicting $Y^{i}$ given $Z^{i}$ and $X^{i}$, while the second term aims to reconstruct $X^{i}$ given $Z^{i}$.



\textbf{Upper bound for} $I(X^{o};Z^{i})$. Next, to minimize the MI between the latent representation $Z^{i}$ and historical series $X^{o}$, 
we adopt the sampled vCLUB as defined in~\cite{cheng2020club}:
\begin{equation}
    I_{v}(X^{o};Z^{i})=\frac{1}{N} \sum_{n=1}^{N} \left [  \log q_{\theta}(z_{n}^{i}|x_{n}^{o})- \log q_{\theta}(z_{n}^{i}|x^{o}_{k'_{n}}) \right ],
\end{equation}
where $(z_{n}^{i},x^{o}_{k'_{n}})$ represents a negative pair and $k_{n}'$ is uniformly selected from indices $\{1,2,...N\}$. 
By minimizing $I_{v}(X^{o};Z^{i})$, we can effectively minimize $I(X^{o}; Z^{i})$. 
It enables the model to extract relevant cross-variable information while eliminating irrelevant information.

Finally, We can convert the intractable objective function $\mathcal{R}_{IB}^{i}$ of all {channels} in Eq.~\ref{ib} as:

\begin{equation}
\begin{aligned}
    \mathcal{L}_{IB}&= \frac{1}{C} \sum_{i=1}^{C}\left[-I_v(X^{i},Y^{i};Z^{i}) + \beta I_{v}(X^{o};Z^{i}) \right] \nonumber\\
    &{\geq -\frac{1}{C}\sum_{i=1}^{C}\mathcal{R}_{IB}^{i}}=-\mathcal{R}_{IB}.
\end{aligned}
\end{equation}
The objective function $\mathcal{L}_{IB}$ is a tractable approximation of the original objective function $\mathcal{R}_{IB}$, we can effectively maximize $\mathcal{R}_{IB}^{i}$ by minimizing $\mathcal{L}_{IB}$.

\subsection{Temporal Correlation Aware Modeling}
\label{approach:TAM}

\begin{figure}[ht]
\vskip 0.2in
\begin{center}
\centerline{\includegraphics[width=\columnwidth]{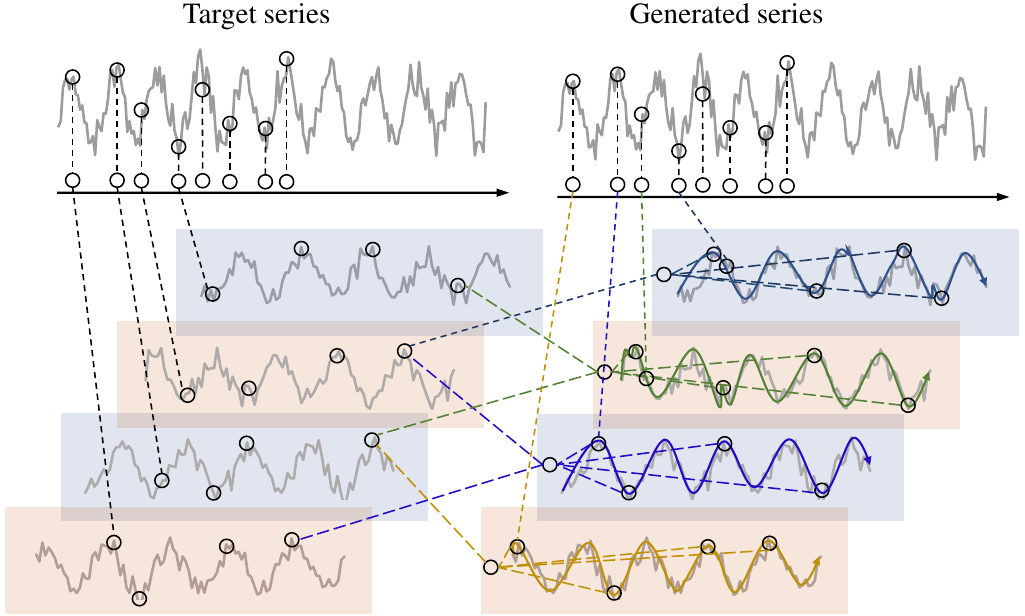}}
\caption{Architecture of \ms \xspace with 4$\times$ dowmsampling. We downsample the target series and forecasted series utilizing single-forecaster into four sub-sequences, respectively. And then we maximize the mutual information between the adjacent sub-sequences of forecasted series and target series.}
\label{fig:ms}
\end{center}
\vskip -0.2in
\end{figure}


Previous works~\citep{nie2022time,zeng2022transformers,zhou2021informer,zhang2023crossformer} have attempted to alleviate error accumulation effects by using a single-step forecaster, which is typically implemented as a fully-connected network, to predict future time series. Unlike auto-regressive forecasters, which consider dependencies between predicted future time steps, the single-step forecaster assumes that the predicted time steps are independent of each other given the historical time series. Consequently, the training objective of the single-step forecaster can be expressed as:
\begin{equation}
p(y^{i}|z^{i},x^{i})=\prod_{j=1}^{P}p(y^{i}_{j}|z^{i},x^{i})
\end{equation}


While the single-step forecaster outperforms the auto-regressive forecaster, it falls short in capturing the temporal correlations among different timesteps in the target series. In contrast to NLP, time series data is considered a low-density information source~\cite{lin2023petformer}. One unique property of time series data is that the temporal relations, such as trends and seasonality, are similar between downsampling adjacent sub-sequences~\cite{liu2022scinet}. Building upon the above observation, we propose {\ms} \xspace to enhance the correlation of predicted future time steps by iteratively down-sampling the time series and optimizing the mutual information between consecutive sub-sequences of both the forecasted and target series. Next, we will introduce \ms \xspace in detail.

{After extracting cross-variable feature $Z^{i}$ as described in Sec~\ref{approach:CAM}, we first generate $\hat{Y^{i}}$ using a single-step forecaster that leverages the historical data of the i-th channel $X^{i}$ and $Z^{i}$ }.
The forecasted series $\hat{Y}$ and target series $Y$ are then downsampled $N$ times. For the n-th ($n \leq N$) downsampling, we generate $m$ sub-sequences $\hat{Y}= \{ \hat{Y}_{1},...,\hat{Y}_{m} \}$,$Y=\{ Y_{1},...,Y_{m} \}$, where $m=2^{n}$ and $\hat{Y}_{j} \in R^{\frac{P}{2^{n}} \times C}$.
Next, we aim to maximize the mutual information $\hat{Y}_{j}^{i}$ and $Y_{j-1}^{i}$, $Y_{j+1}^{i}$, given $X^{i}$, where $1<j<m$. This is achieved by calculating $\mathcal{L}_{n}$ for the $n$-the downsampling:
\begin{equation}
\begin{aligned}
    \mathcal{L}_{n}&=-\frac{1}{mC} \sum_{i=1}^{C} \Big[  I(Y_{2}^{i};\hat{Y}_{1}^{i}|X^{i})+I(Y_{m-1}^{i};\hat{Y}_{m}^{i}|X^{i}) \nonumber \\
    &+ \sum_{j=2}^{m-1} I(Y_{j-1}^{i};\hat{Y}_{j}^{i}|X^{i}) +I(Y_{j+1}^{i};\hat{Y}_{j}^{i}|X^{i})  \Big].
\end{aligned}
\end{equation}

We also introduce the variational lower bound for $I(Y^{i}_{j-1};\hat{Y}_{j}^{i}|X^{i})$, which is as follows (More details are shown in Appendix~\ref{sec:a.3}):
\begin{equation}
    I(Y^{i}_{j-1};\hat{Y}_{j}^{i}|X^{i}) \geq \E_{p(y_{j-1}^{i},\hat{y}_{j}^{i},x^{i})} \left[ p_{\theta}(y_{j-1}^{i}|\hat{y}_{j}^{i},x^{i}) \right].
\end{equation}

Furthermore, considering the efficiency, we make the assumption that the time steps of a sub-sequence are independent given the adjacent sub-sequence. 
As a result, we can simplify the mutual information term, $I(Y^{i}_{j-1};\hat{Y}_{j}^{i}|X^{i})$ as $I(Y^{i}_{j-1};\hat{Y}_{j}^{i}|X^{i})= \sum_{k=1}^{\frac{P}{2^{n}}} \left[ I(Y^{i}_{j-1,k};\hat{Y}_{j}^{i}|X^{i}) \right]$. This allows us to generate the entire sub-sequence in a single step without requiring auto-regression.

For the n-th downsampling, \ms \xspace will generate $2\times(2^{n}-1)$ sub-sequences denoted as $\hat{Y}'=\{ \hat{Y}_{1}^{r}, \hat{Y}_{2}^{l},\hat{Y}_{2}^{r},..., \hat{Y}_{m}^{l}\}$.
These sub-sequences, except for the ones and the ends, are predicted by their left and right adjacent sub-sequences respectively. We then splice these $2\times(2^{n}-1)$ sub-sequences into a new series $\hat{Y}_{n}=\{ \hat{Y}_{1}^{r}, \frac{\hat{Y}_{2}^{l}+\hat{Y}_{2}^{r}}{2},...,\hat{Y}_{m}^{l}\}$. After $N$ downsamplings, we have generated $N+1$ series. We use these $N+1$ series as the final forecasting results, resulting in the following loss function:

\begin{equation}
    \mathcal{L}_{p}= ||Y-(\lambda \sum_{n=1}^{N}{\frac{\hat{Y}_{n}}{N}}+(1-\lambda)\hat{Y})||_{2}^{2}
\end{equation}


{In contrast to single-step forecasters that generate multi-step predictions without considering the correlation between the predicted series, our proposed method, referred to as \ms, 
explicitly models the correlation of predicted future time steps and allows to generate more accurate and coherent predictions. 
}

Integrating \mf \xspace and \ms, the total loss of \framework \xspace  can be written as :
\begin{equation}
    \mathcal{L}_{total}=\mathcal{L}_{IB} + \sum_{n=1}^{N} \mathcal{L}_{n}+\mathcal{L}_{p}.
\end{equation}




\section{Experiments}
\begin{figure*}[h!]
	\centering
	\subfloat[Informer on ETTh1]{
		\begin{minipage}[t]{0.25\linewidth}
			\centering
			\includegraphics[width=1.54in]{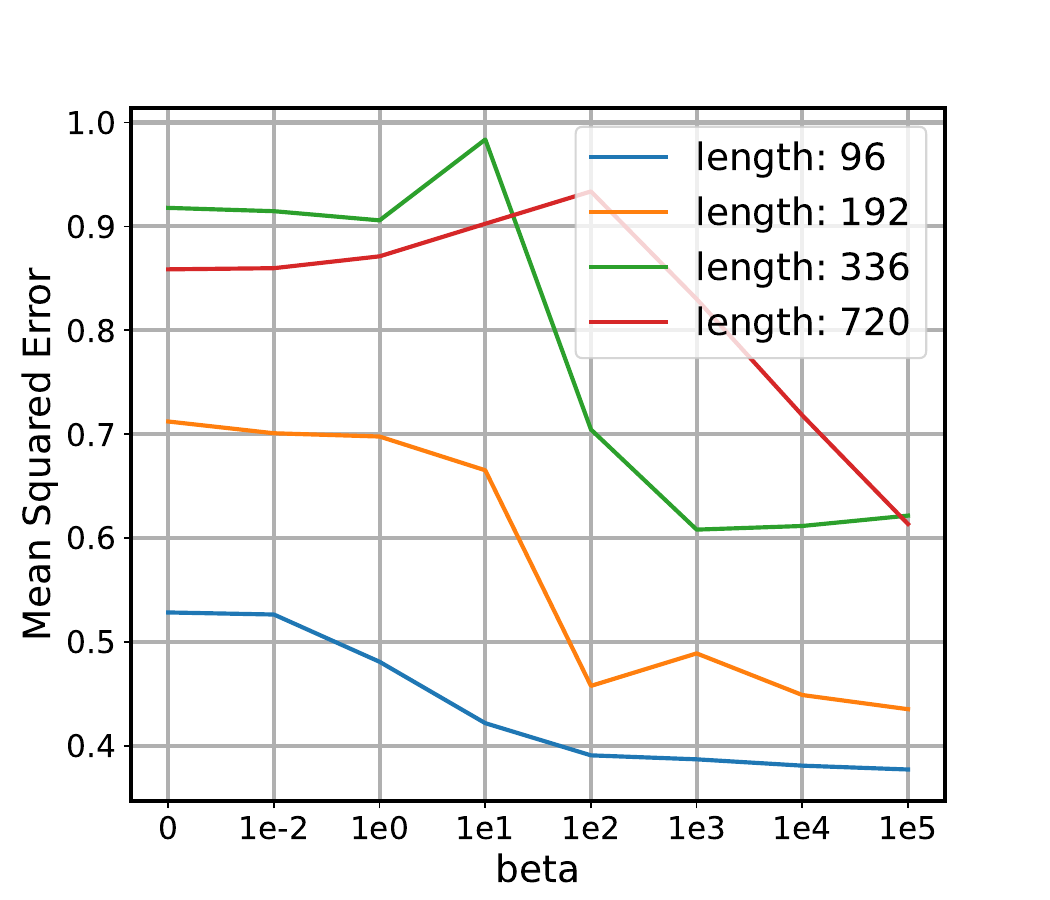}\\
			\vspace{0.01cm}
		\end{minipage}%
	}%
	\subfloat[Stationary on ETTh1]{
		\begin{minipage}[t]{0.25\linewidth}
			\centering
			\includegraphics[width=1.5in]{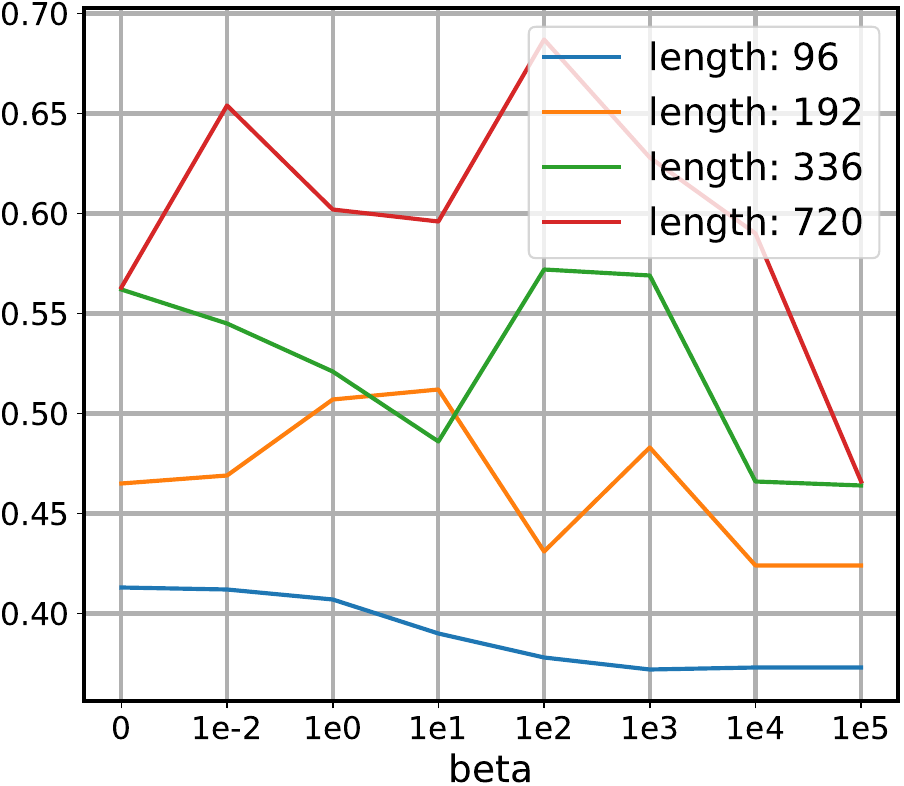}\\
			\vspace{0.01cm}
		\end{minipage}%
	}%
	\subfloat[PatchTST on ETTh1]{
		\begin{minipage}[t]{0.25\linewidth}
			\centering
			\includegraphics[width=1.5in]{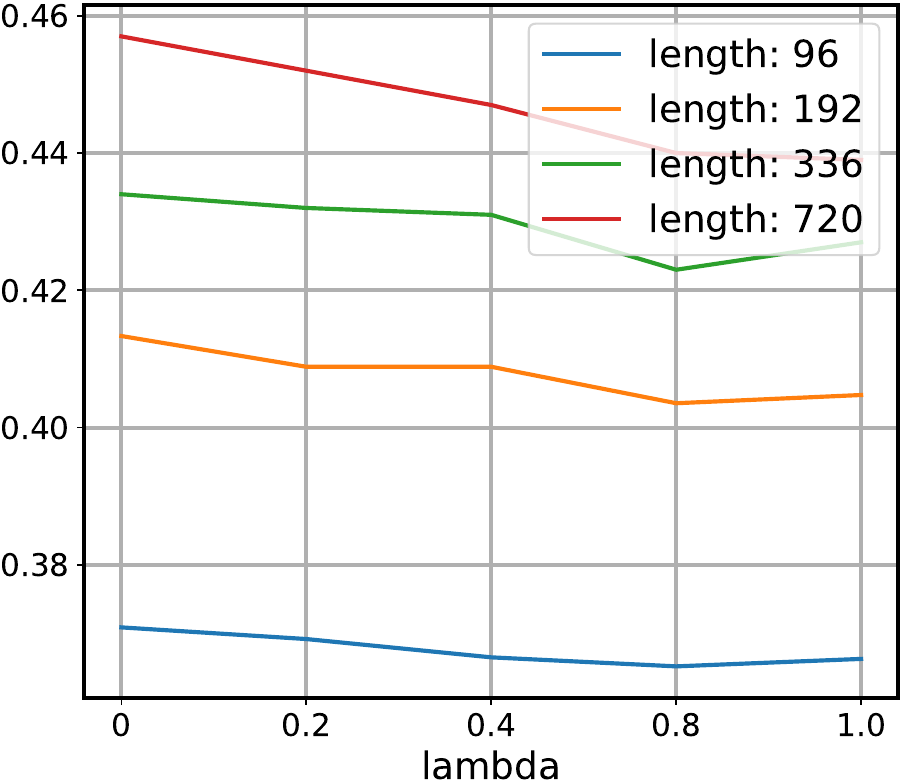}\\
			\vspace{0.01cm}
		\end{minipage}%
	}%
 \subfloat[RMLP on ETTh1]{
		\begin{minipage}[t]{0.25\linewidth}
			\centering
			\includegraphics[width=1.5in]{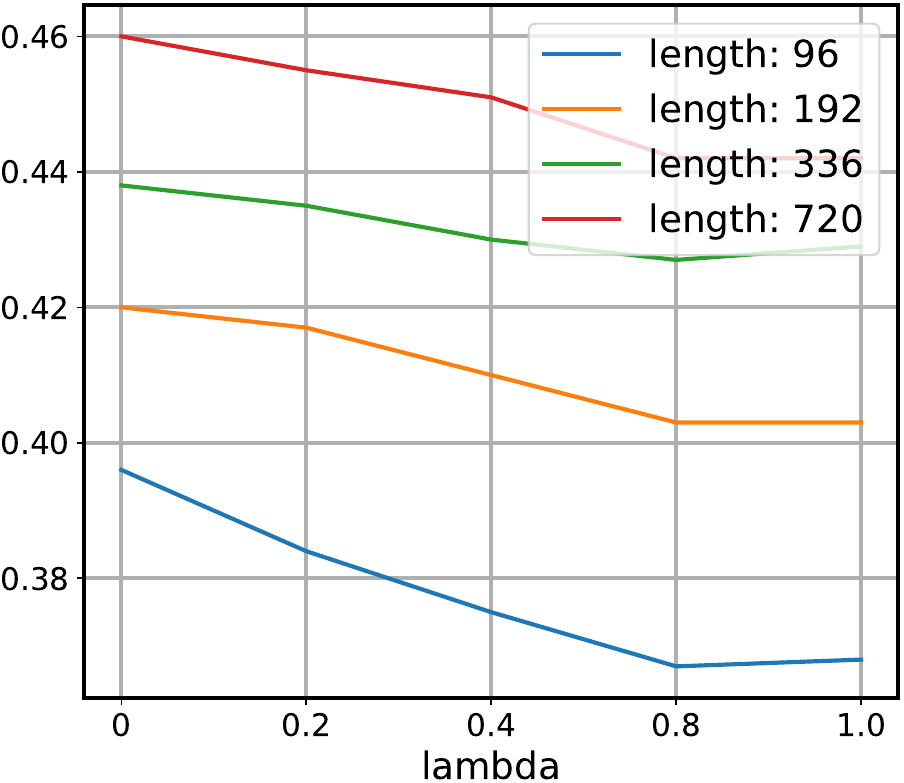}\\
			\vspace{0.01cm}
		\end{minipage}%
	}%
	\centering
	\caption{{Evaluation on hyper-parameter $\beta$ and $\lambda$. We evaluate the impact of $\beta$ with Informer and Stationary on the ETTh1 dataset, we also evaluate $\lambda$ with PatchTST and RMLP on the ETTh1 dataset.}}
	\vspace{-0.2cm}
	\label{fig:ablation_beta}
\end{figure*}
\begin{table*}[h!]\small
\caption{Multivariate long-term series forecasting results on Channel-mixing models with different prediction lengths $O \in \{96,192,336,720\}$. We set the input length $I$ as 96 for all the models. The best result is indicated in bold font. {\textit{Avg} is averaged from all four prediction lengths.
and \textit{Pro} means the relative MSE or MAE reduction. 
(See Table~\ref{tab:ablation_dependent} for the full results.)}}
    \centering
    \tabcolsep=0.15cm
    \begin{tabular}{c|c|cccc|cccc|cccc}
    
    \hline
    \multicolumn{2}{c|}{\multirow{2}{*}{Models}} & \multicolumn{4}{c|}{Informer} & \multicolumn{4}{c|}{Stationary} & \multicolumn{4}{c}{Crossformer} \\
    
    \multicolumn{2}{c|}{} & \multicolumn{2}{c}{Original} & \multicolumn{2}{c|}{w/Ours} & \multicolumn{2}{c}{Original} & \multicolumn{2}{c|}{w/Ours} & \multicolumn{2}{c}{Original} & \multicolumn{2}{c}{w/Ours} \\
    \hline
    \multicolumn{2}{c|}{Metric} & MSE & MAE & MSE & MAE & MSE & MAE & MSE & MAE & MSE & MAE & MSE & MAE \\
    \hline
    \multirow{4}{*}{\rotatebox{90}{ETTm2}} 
    & 96 & 0.365 & 0.453 & \textbf{0.187} & \textbf{0.282} & 0.201 & 0.291 & \textbf{0.175} & \textbf{0.256} & 0.281 & 0.373 & \textbf{0.186} & \textbf{0.281} \\
    & 192 & 0.533 & 0.563 & \textbf{0.277} & \textbf{0.351} & 0.275 & 0.335 & \textbf{0.238} & \textbf{0.297} & 0.549 & 0.520 & \textbf{0.269} & \textbf{0.341} \\
    & 336 & 1.363 & 0.887 & \textbf{0.380} & \textbf{0.420} & 0.350 & 0.377 & \textbf{0.299} & \textbf{0.336} & 0.729 & 0.603 & \textbf{0.356} & \textbf{0.396} \\
    & 720 & 3.379 & 1.338 & \textbf{0.607} & \textbf{0.549} & 0.460 & 0.435 & \textbf{0.398} & \textbf{0.393} & 1.059 & 0.741 & \textbf{0.493} & \textbf{0.482} \\
    \hline
    & Avg & 1.410 & 0.810 & \textbf{0.362} & \textbf{0.400} & 0.321 & 0.359 & \textbf{0.277} & \textbf{0.320} & 0.654 & 0.559 & \textbf{0.326} & \textbf{0.375} \\
    \hline 
    & Pro & - & - & \textbf{74.3\%} & \textbf{50.6\%} & - & - & \textbf{13.7\%} & \textbf{10.8\%} & - & - & \textbf{50.1\%} & \textbf{32.9\%} \\
    \hline
    \multirow{4}{*}{\rotatebox{90}{Weather}} 
    & 96 & 0.300 & 0.384 & \textbf{0.179} & \textbf{0.249} & 0.181 & 0.230 & \textbf{0.166} & \textbf{0.213} & 0.158 & 0.236 & \textbf{0.149} & \textbf{0.218} \\
    & 192 & 0.598 & 0.544 & \textbf{0.226} & \textbf{0.296} & 0.286 & 0.312 & \textbf{0.218} & \textbf{0.260} & 0.209 & 0.285 & \textbf{0.202} & \textbf{0.272} \\
    & 336 & 0.578 & 0.523 & \textbf{0.276} & \textbf{0.334} & 0.319 & 0.335 & \textbf{0.274} & \textbf{0.300} & 0.265 & 0.328 & \textbf{0.256} & \textbf{0.313} \\
    & 720 & 1.059 & 0.741 & \textbf{0.332} & \textbf{0.372} & 0.411 & 0.393 & \textbf{0.351} & \textbf{0.353} & 0.376 & 0.397 & \textbf{0.329} & \textbf{0.366} \\
    \hline
    & Avg & 0.633 & 0.548 & \textbf{0.253} & \textbf{0.312} & 0.299 & 0.317 & \textbf{0.252} & \textbf{0.281} & 0.252 & 0.311 & \textbf{0.234} & \textbf{0.292} \\
    \hline 
    & Pro & - & - & \textbf{60.0\%} & \textbf{43.0\%} & - & - & \textbf{15.7\%} & \textbf{11.3\%} & - & - & \textbf{7.1\%} & \textbf{6.1\%} \\
    \hline
    \multirow{4}{*}{\rotatebox{90}{Traffic}} 
    & 96 & 0.719 & 0.391 & \textbf{0.505} & \textbf{0.348} & 0.599 & 0.332 & \textbf{0.459} & \textbf{0.311} & 0.609 & 0.362 & \textbf{0.529} & \textbf{0.334} \\
    & 192 & 0.696 & 0.379 & \textbf{0.521} & \textbf{0.354} & 0.619 & 0.341 & \textbf{0.475} & \textbf{0.315} & 0.623 & 0.365 & \textbf{0.519} & \textbf{0.327} \\
    & 336 & 0.777 & 0.420 & \textbf{0.520} & \textbf{0.337} & 0.651 & 0.347 & \textbf{0.486} & \textbf{0.319} & 0.649 & 0.370 & \textbf{0.521} & \textbf{0.337} \\
    & 720 & 0.864 & 0.472 & \textbf{0.552} & \textbf{0.352} & 0.658 & 0.358 & \textbf{0.522} & \textbf{0.338} & 0.758 & 0.418 & \textbf{0.556} & \textbf{0.350} \\
    \hline
    & Avg & 0.764 & 0.415 & \textbf{0.524} & \textbf{0.347} & 0.631 & 0.344 & \textbf{0.485} & \textbf{0.320} & 0.659 & 0.378 & \textbf{0.531} & \textbf{0.337} \\
    \hline
    & Pro & - & - & \textbf{31.4\%} & \textbf{16.3\%} & - & - & \textbf{23.1\%} & \textbf{6.9\%} & - & - & \textbf{19.4\%} & \textbf{10.8\%} \\
    \hline
    \multirow{4}{*}{\rotatebox{90}{Electricity}} 
    & 96 & 0.274 & 0.368 & \textbf{0.195} & \textbf{0.300} & 0.168 & 0.271 & \textbf{0.154} & \textbf{0.256} & 0.170 & 0.279 & \textbf{0.150} & \textbf{0.248} \\
    & 192 & 0.296 & 0.386 & \textbf{0.193} & \textbf{0.291} & 0.186 & 0.285 & \textbf{0.163} & \textbf{0.263} & 0.198 & 0.303 & \textbf{0.168} & \textbf{0.263} \\
    & 336 & 0.300 & 0.394 & \textbf{0.206} & \textbf{0.300} & 0.194 & 0.297 & \textbf{0.178} & \textbf{0.279} & 0.235 & 0.328 & \textbf{0.200} & \textbf{0.290} \\
    & 720 & 0.373 & 0.439 & \textbf{0.241} & \textbf{0.332} & 0.224 & 0.316 & \textbf{0.201} & \textbf{0.299} & 0.270 & 0.360 & \textbf{0.235} & \textbf{0.323} \\
    \hline
    & Avg & 0.310 & 0.397 & \textbf{0.208} & \textbf{0.305} & 0.193 & 0.292 &  0.174 & \textbf{0.274} & 0.218 & 0.317 & \textbf{0.188} & \textbf{0.281} \\

    \hline 
    & Pro & - & - & \textbf{32.9\%} & \textbf{23.1\%} & - & - & \textbf{9.8\%}  & \textbf{6.1\%} & - & - & \textbf{13.7\%} & \textbf{11.3\%} \\
    \hline
    \end{tabular}
    \label{tab:main_results_channel_dependent}
\end{table*}
In this section, we extensively evaluate the proposed \framework \xspace on 10 real-world benchmarks using various Channel-mixing and Channel-Independence models,
including state-of-the-art models.

\textbf{Baselines.} In order to demonstrate the versatility of our proposed method, \framework, we evaluate it on various deep-learning-based forecasting models, including several popular baselines as well as the state-of-the-art method.  
For Channel-mixing models, we select Informer~\citep{zhou2021informer}, Non-stationary Transformer~\citep{liu2022non}, denoting as Stationary, and Crossformer~\citep{zhang2023crossformer}. As for Channel-Independence models, we utilize PatchTST~\citep{nie2022time} and also introduce RMLP which consists of two linear layers with ReLU activation and also incorporates reversible instance normalization~\citep{kim2021reversible}. (see Appendix~\ref{baselines} for more details) 

\textbf{Datasets.} We extensively include 10 real world datasets in our experiments, including four ETT datasets (ETTh1, ETTh2, ETTm1, ETTm2)~\cite{zhou2021informer}. Electricity, Weather, Traffic~\cite{chen2021autoformer}, and PEMS covering energy, transportation and weather domains (See Appendix~\ref{sec:dataset_details} for more details on the datasets). We follow the standard protocol that divides each dataset into the training, validation, and testing subsets according to the chronological order. The split ratio is 6:2:2 for the ETT dataset and 7:1:2 for others.



\subsection{Main Results}

Table~\ref{tab:main_results_channel_dependent} compares the forecasting accuracy of the Channel-mixing baselines and \framework.
The results clearly demonstrate that \framework \xspace consistently outperforms all three baselines, namely Informer, Stationary, and Crossformer, by a significant margin. This improvement is particularly notable for long sequence predictions, which are generally more challenging and prone to the model relying on irrelevant cross-variable information. \framework \xspace exhibits stable performance, whereas the baselines experience a substantial increase in forecasting error as the prediction length is extended.
For example, on the ETTm2 dataset, when the prediction length increases from 96 to 720, the forecasting error of Informer rises significantly from 0.365 to 3.379. In contrast, \framework \xspace shows a much slight increase in error.
A similar tendency appears with the other prediction lengths, datasets, and baseline models as well. These results demonstrate that \framework \xspace makes the baseline models more robust to prediction target series. 

\begin{figure*}[h!]
	\centering
	\subfloat[Test error on Informer]{
		\begin{minipage}[t]{0.3\linewidth}
			\centering
			\includegraphics[width=1.76in]{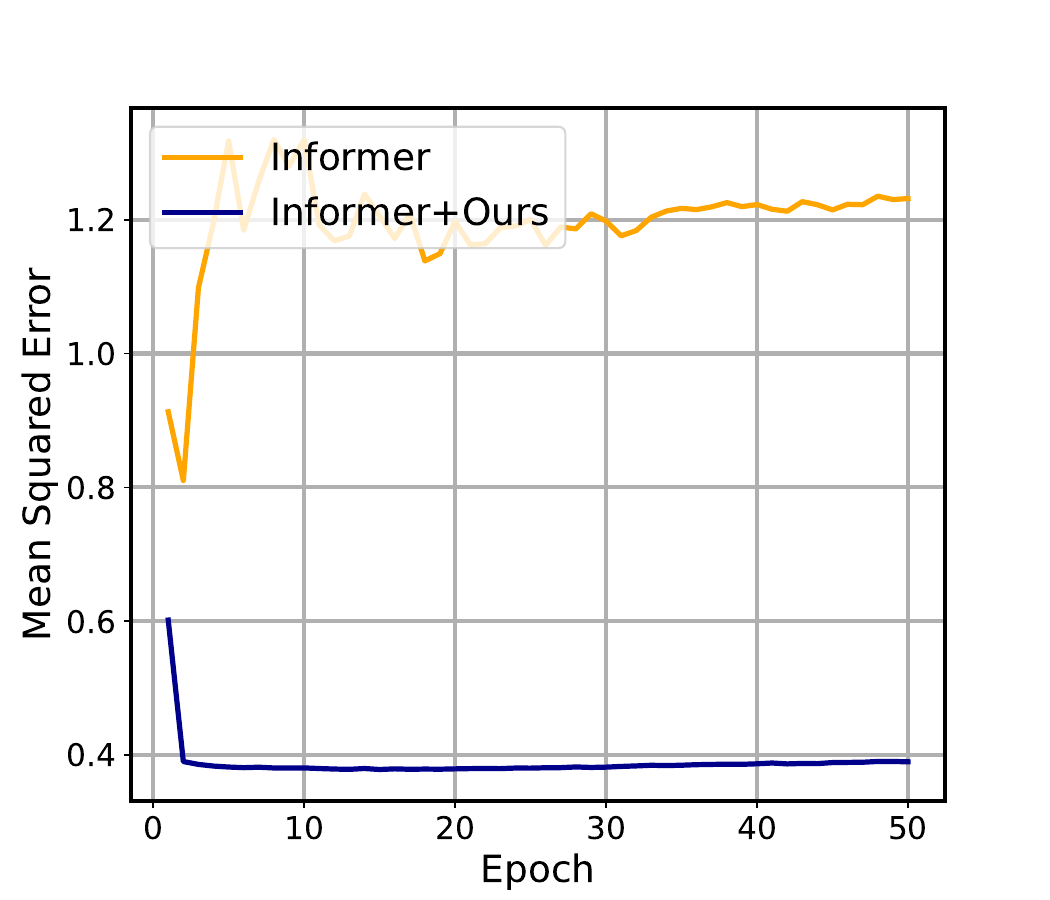}\\
			\vspace{0.015cm}
		\end{minipage}%
	}%
	\subfloat[Test error on Stationary]{
		\begin{minipage}[t]{0.3\linewidth}
			\centering
			\includegraphics[width=1.8in]{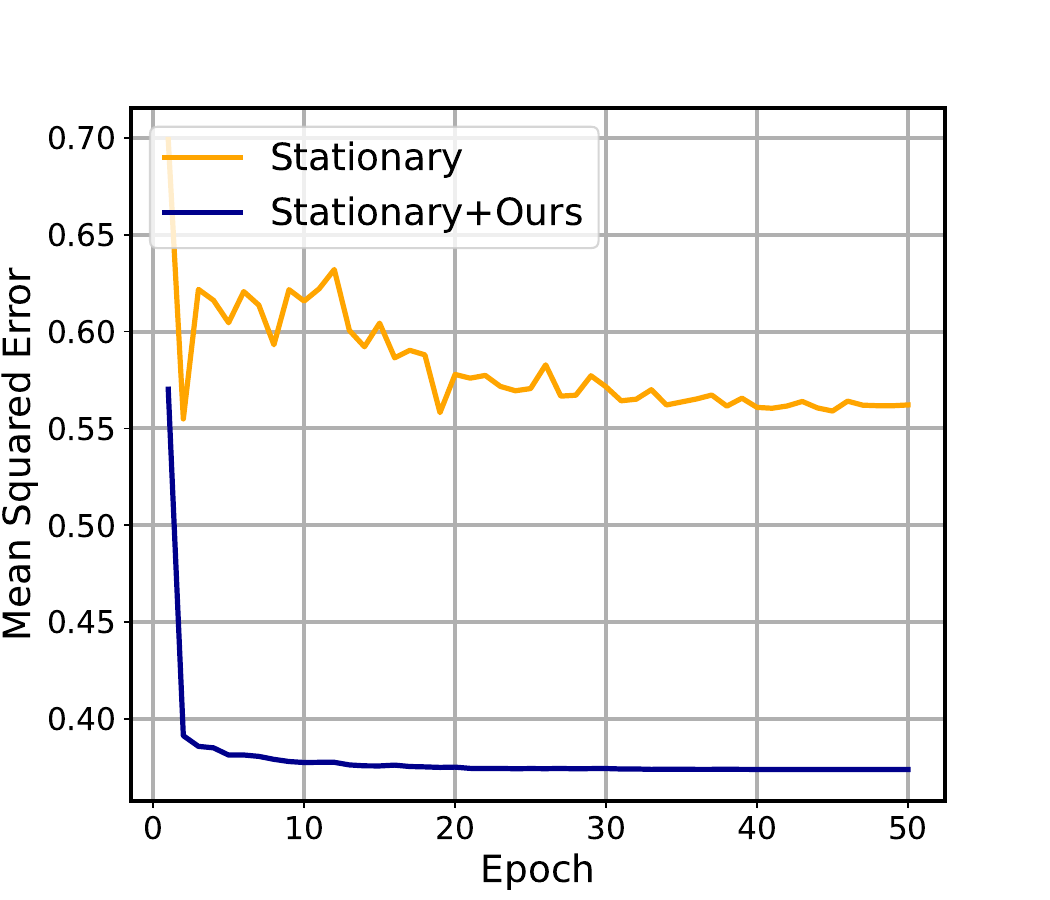}\\
			\vspace{0.02cm}
		\end{minipage}%
	}%
	\subfloat[Test error on Crossformer]{
		\begin{minipage}[t]{0.3\linewidth}
			\centering
			\includegraphics[width=1.8in]{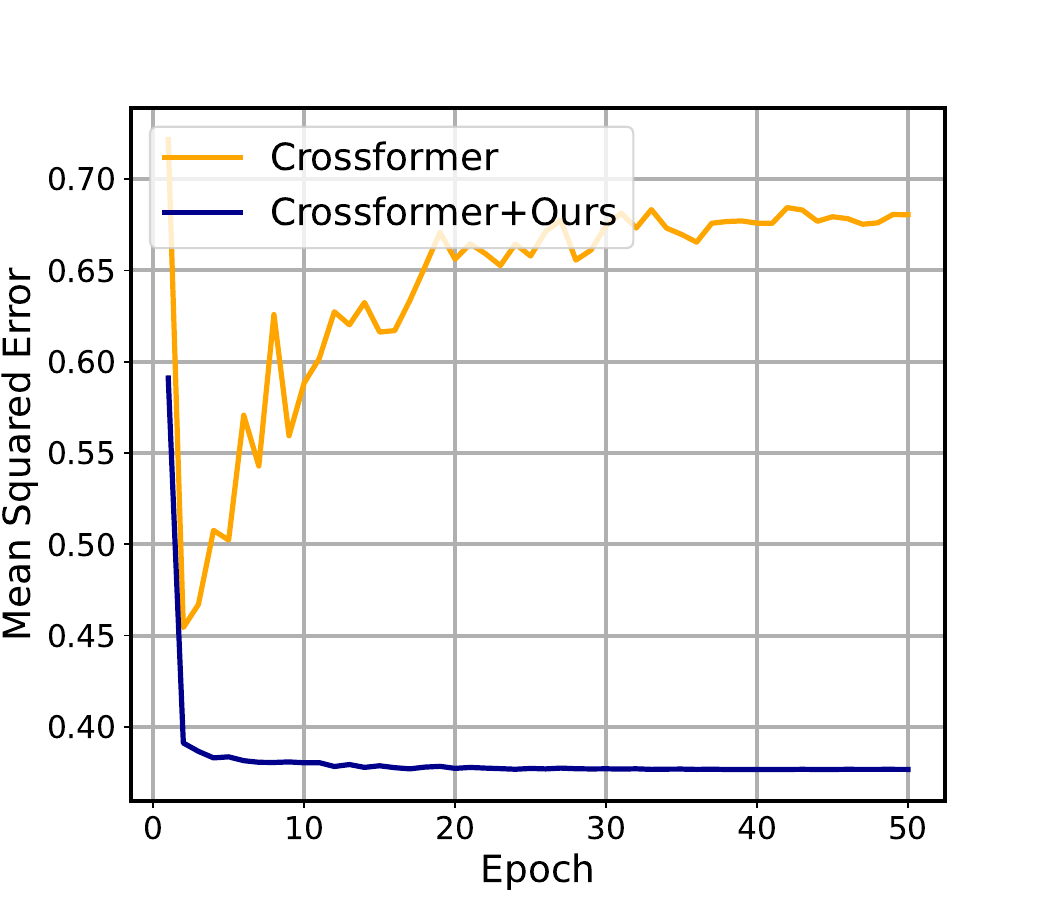}\\
			\vspace{0.02cm}
		\end{minipage}%
	}%
	\centering
	\caption{\textbf{Test error for each training epoch}. We train the baselines and integrated with our \framework \xspace for 50 epochs on ETTh1 with the history and prediction length are both 96.}
	\vspace{-0.2cm}
	\label{fig:test_loss}
\end{figure*}

To gain further insights into why \framework \xspace outperforms baselines, we visualize the testing error for each epoch in Figure~\ref{fig:test_loss}. Overall, \framework \xspace exhibits lower and more stable test errors compared to the baselines. Moreover, the baselines are highly prone to overfitting in the early stages of training, whereas \framework \xspace effectively mitigates this issue.

\begin{table*}[h!]
    \centering
    \caption{Multivariate long-term series forecasting results on Channel-Independence models with different prediction lengths. We set the input length $I$ as 336 for all the models. The best result is indicated in bold font. See {Table~\ref{tab:main_results_channel_independent_full}} in the Appendix for the full results.}
    \tabcolsep=0.08cm
    \begin{tabular}{ccc|cccc|cccc|cccc}
    \hline
         \multicolumn{2}{c}{\multirow{2}{*}{Models}} & \multirow{2}{*}{Metric} & \multicolumn{4}{c|}{ETTh1} & \multicolumn{4}{c|}{ETTm1} & \multicolumn{4}{c}{Traffic} \\
         \multicolumn{2}{c}{} & & 96 & 192 & 336 & 720 & 96 & 192 & 336 & 720  & 96 & 192 & 336 & 720 \\
         \hline
         \multirow{4}{*}{\rotatebox{90}{PatchTST}} & \multirow{2}{*}{Original} & MSE & 0.375 & 0.414 & 0.440 & 0.460 & 0.290 & 0.332 & 0.366 & 0.420 & 0.367 & 0.385 & 0.398 & 0.434 \\
          & & MAE &  0.399 & 0.421 & 0.440 & 0.473 & 0.342 & 0.369 & 0.392 & 0.424 & 0.251 & 0.259 & 0.265 & 0.287 \\
          & \multirow{2}{*}{\textbf{w/Ours}} & MSE & \textbf{0.365} & \textbf{0.403} & \textbf{0.427} & \textbf{0.433} & \textbf{0.283} & \textbf{0.322} & \textbf{0.356} & \textbf{0.407} & \textbf{0.358} & \textbf{0.379} & \textbf{0.391} & \textbf{0.425} \\
          & & MAE  & \textbf{0.389} & \textbf{0.413} & \textbf{0.428} & \textbf{0.453} & \textbf{0.335} & \textbf{0.359} & \textbf{0.382} & \textbf{0.417} & \textbf{0.245} & \textbf{0.254} & \textbf{0.261} & \textbf{0.280} \\ 
          \hline
          \multirow{4}{*}{\rotatebox{90}{RMLP}} & \multirow{2}{*}{Original} & MSE & 0.380 & 0.414 & 0.439 & 0.470 & 0.290 & 0.329 & 0.364 & 0.430 & 0.383 & 0.401 & 0.414 & 0.443 \\
          & & MAE  & 0.401 & 0.421 & 0.436 & 0.471 & 0.343 & 0.368 & 0.390 & 0.426 & 0.269 & 0.276 & 0.282 & 0.309 \\
          & \multirow{2}{*}{\textbf{w/Ours}} & MSE & \textbf{0.367} & \textbf{0.404} & \textbf{0.426} & \textbf{0.439} & \textbf{0.285} & \textbf{0.322} & \textbf{0.358} & \textbf{0.414} & \textbf{0.364} & \textbf{0.384} & \textbf{0.398} & \textbf{0.428} \\
          & & MAE & \textbf{0.391} & \textbf{0.413} & \textbf{0.429} & \textbf{0.459} & \textbf{0.335} & \textbf{0.359} & \textbf{0.381} & \textbf{0.413} & \textbf{0.249} & \textbf{0.258} & \textbf{0.266} & \textbf{0.284} \\
         \hline
    \end{tabular}
    \label{tab:marin_results_channel_independent}
\end{table*}
We also present the forecasting results of Channel-independence baselines in Table~\ref{tab:marin_results_channel_independent}. It is worth noting that \framework \xspace also surpasses Channel-independence baselines, indicating that although Channel-independence models exhibit promising results, incorporating cross-variable features can further enhance their effectiveness. Furthermore, we evaluate \framework \xspace on the PEMS datasets, which consist of variables that exhibit clear geographical correlations. The results in Table~\ref{tab:pems_results} reveal a significant performance gap between PatchTST and RMLP compared to Informer, suggesting that Channel-independence models may not be optimal in scenarios where there are clear correlations between variables. In contrast, our framework exhibits improved performance for both Channel-mixing and Channel-independence models. 

\begin{table*}[!]
    \centering
    \caption{Multivariate long-term series forecasting results on baselines and PEMS datasets with different prediction lengths. We set the input length $I$ as 336 for all the models. The best result is indicated in bold font. (See {Table~\ref{tab:pems_ablation_results}} for the full results.) }
    \tabcolsep=0.08cm
    \begin{tabular}{ccc|cccc|cccc|cccc}
    \hline
     \multicolumn{2}{c}{\multirow{2}{*}{Models}} & \multirow{2}{*}{Metric} & \multicolumn{4}{c|}{PEMS03} & \multicolumn{4}{c|}{PEMS04} & \multicolumn{4}{c}{PEMS08} \\
     \multicolumn{2}{c}{} & & 96 & 192 & 336 & 720 & 96 & 192 & 336 & 720 & 96 & 192 & 336 & 720 \\
     \hline
     \multirow{4}{*}{\rotatebox{90}{\footnotesize PatchTST}} & \multirow{2}{*}{Original} & MSE & 0.180 & 0.207 & 0.223 & 0.291 & 0.195 & 0.218 & 0.237 & 0.321 & 0.239 & 0.292 & 0.314 & 0.372 \\
          & & MAE &  0.281 & 0.295 & 0.309 & 0.364 & 0.296 & 0.314 & 0.329 & 0.394 & 0.324 & 0.351 & 0.374 & 0.425 \\
          & \multirow{2}{*}{\textbf{w/Ours}} & MSE & 
          \textbf{0.115} & \textbf{0.154} & \textbf{0.164} & \textbf{0.198} & \textbf{0.110} & \textbf{0.118} & \textbf{0.129} & \textbf{0.149} & \textbf{0.114} & \textbf{0.160} & \textbf{0.177} & \textbf{0.209}\\
          & & MAE & \textbf{0.223} & \textbf{0.251} & \textbf{0.256} & \textbf{0.286} & \textbf{0.221} & \textbf{0.224} & \textbf{0.237} & \textbf{0.261} & \textbf{0.218} & \textbf{0.243} & \textbf{0.241} & \textbf{0.281} \\ 
          \hline
          \multirow{4}{*}{\rotatebox{90}{\footnotesize RMLP}} & \multirow{2}{*}{Original} & MSE & 0.160 & 0.184 & 0.201 & 0.254 & 0.175 & 0.199 & 0.210 & 0.255 & 0.194 & 0.251 & 0.274 & 0.306 \\
          & & MAE  & 0.257 & 0.277 & 0.291 & 0.337 & 0.278 & 0.294 & 0.306 & 0.348 & 0.279 & 0.311 & 0.328 & 0.365 \\
          & \multirow{2}{*}{\textbf{w/Ours}} & MSE & \textbf{0.117} & \textbf{0.159} & \textbf{0.146} & \textbf{0.204} & \textbf{0.103} & \textbf{0.114} & \textbf{0.130} & \textbf{0.154} & \textbf{0.116} & \textbf{0.156} & \textbf{0.175} & \textbf{0.181} \\
          & & MAE & \textbf{0.228} & \textbf{0.252} & \textbf{0.246} & \textbf{0.285} & \textbf{0.211} & \textbf{0.219} & \textbf{0.236} & \textbf{0.264} & \textbf{0.215} & \textbf{0.235} & \textbf{0.242} & \textbf{0.255} \\
          \hline

    \end{tabular}
    \label{tab:pems_results}
\end{table*}

\subsection{Ablation Study}
\begin{table*}[!]\normalsize
\centering
\tabcolsep=0.15cm
\caption{Component ablation of \framework. We set the input length $I$ as 336 for PatchTST and 96 for Informer. The best results are in \textbf{bold} and the second best are \underline{underlined}. (See {Table~\ref{tab:ablation_dependent}} and {Table~\ref{tab:ablation_independent}} in the Appendix for the full ablation results.)}
\begin{tabular}{cc|cccccc|cccccc}
\hline
     \multicolumn{2}{c|}{\multirow{2}{*}{Models}} & \multicolumn{6}{c|}{Informer} & \multicolumn{6}{c}{PatchTST} \\
     \multicolumn{2}{c|}{} & \multicolumn{2}{c}{Original} & \multicolumn{2}{c}{+\ms} & \multicolumn{2}{c|}{+\framework} & \multicolumn{2}{c}{Original} & \multicolumn{2}{c}{+\ms} & \multicolumn{2}{c}{+\framework} \\
     \hline
     \multicolumn{2}{c|}{Metric} & MSE & MAE & MSE & MAE & MSE & MAE & MSE & MAE & MSE & MAE & MSE & MAE \\
     \hline 
       
       \multirow{4}{*}{\rotatebox{90}{ETTm1}} 
       & 96 & 0.672 & 0.571 & \underline{0.435} & \underline{0.444} & \textbf{0.335} & \textbf{0.369} & 0.290 & 0.342 & \underline{0.286} & \underline{0.337} & \textbf{0.283} & \textbf{0.335} \\
      & 192 & 0.795 & 0.669 & \underline{0.473} & \underline{0.467} & \textbf{0.380} & \textbf{0.390} & 0.332 & 0.369 & \underline{0.326} & \underline{0.366} & \textbf{0.322} & \textbf{0.363} \\
      & 336 & 1.212 & 0.871 & \underline{0.545} & \underline{0.518} & \textbf{0.423} & \textbf{0.426} & 0.366 & 0.392 & \underline{0.359} & \underline{0.388} & \textbf{0.356} & \textbf{0.385} \\
      & 720 & 1.166 & 0.823 & \underline{0.669} & \underline{0.589} & \textbf{0.505} & \textbf{0.480} & 0.420 & 0.424 & \underline{0.408} & \textbf{0.413} & \textbf{0.407} & \underline{0.417} \\
      \hline
       \multirow{4}{*}{\rotatebox{90}{Weather}} 
       &  96 & 0.300 & 0.384 & \underline{0.277} & \underline{0.354} & \textbf{0.179}& \textbf{0.249} & 0.152 & 0.199 & \underline{0.149} & \underline{0.197} & \textbf{0.144} & \textbf{0.194} \\
        &192 & 0.598 & 0.544 & \underline{0.407} & \underline{0.447} & \textbf{0.226} & \textbf{0.296} & 0.197 & 0.243 & \underline{0.192} & \textbf{0.238} & \textbf{0.189} & \textbf{0.238} \\ 
        &336 & 0.578 & 0.523 & \underline{0.529} & \underline{0.520}& \textbf{0.276} & \textbf{0.334} & 0.250 & 0.284 & \underline{0.247}& \underline{0.280} & \textbf{0.239} & \textbf{0.279} \\ 
       &720 & 1.059 & 0.741 & \underline{0.951} & \underline{0.734} & \textbf{0.332} & \textbf{0.372} & \underline{0.320} & 0.335 & 0.321 & \underline{0.332} & \textbf{0.312} & \textbf{0.331} \\  
       \hline
      \label{tab:ablation}
\end{tabular}
\end{table*}
In our approach, there are two components: \mf \xspace and \ms. To investigate the effectiveness of these components, we conduct an ablation study on the ETTm1 and Weather datasets with Informer and PatchTST. \textbf{+\ms} means that we add \ms \xspace to these baselines and \textbf{+\framework} means that we add both \mf \xspace and \ms \xspace to baselines. 
The results of this ablation study are summarized in Table~\ref{tab:ablation}.
Compared to baselines using the single-step forecaster, \ms \xspace performs better in most settings, which indicates the importance of cross-time correlation. For Channel-mixing models, we find that \framework \xspace can significantly enhance their performance and effectively alleviate the issue of overfitting. For Channel-Independence models, \framework \xspace still improves their performance, which indicates that correctly establishing the dependency between variables is an effective way to improve performance.

\subsection{Effect of Hyper-Parameters}

We investigate the impact of the hyper-parameter $\beta$ on the ETTh1 and ETTm2 datasets with two baselines. In Figure~\ref{fig:ablation_beta} (a) (b),
we vary the value of $\beta$ from $0$ to $1e^{5}$ and evaluate the MSE with different prediction windows for both datasets and baselines.
We find that when $\beta$ is small, baselines exhibit poor and unstable performance. As $\beta$ increases, baselines demonstrate improved and more stable performance. Furthermore, as the prediction window increases, the overfitting problem s become more severe for baselines. 
Therefore, a larger $\beta$ is needed to remove superfluous information and mitigate overfitting. 
Additionally, we evaluate $\lambda$ with PatchTST and RMLP. We observe that a larger $\lambda$ leads to better performance for these models. Moreover, when $\lambda \geq 0.8$, the performance stabilizes. 


\section{Conclusion}
This paper focus on two key factors in MTSF: temporal correlation and cross-variable correlation. To utilize the cross-variable correlation while eliminating the superfluous information, we introduce \textbf{C}ross-Variable Decorrelation \textbf{A}ware \textbf{M}odeling (\mf). In addition, we also propose \textbf{T}emporal Correlation \textbf{A}ware \textbf{M}odeling (\ms) to model temporal correlations of predicted series. Integrating \mf \xspace and \ms, we build a novel time series modeling framework for MTSF termed \framework . Extensive experiments on various real-world MTSF datasets demonstrate the effectiveness of our framework.

 \section*{Broader Impacts}
 Our research is dedicated to innovating time series forecasting techniques to push the boundaries of time series analysis further. While our primary objective is to enhance predictive accuracy and efficiency, we are also mindful of the broader ethical considerations that accompany technological progress in this area. While immediate societal impacts may not be apparent, we acknowledge the importance of ongoing vigilance regarding the ethical use of these advancements. It is essential to continuously assess and address potential implications to ensure responsible development and application in various sectors.

\nocite{langley00}

\bibliography{example_paper}
\bibliographystyle{icml2024}

\newpage
\appendix
\onecolumn
\section{Appendix.}
\subsection{Datasets}
\label{sec:dataset_details}
we use the most popular multivariate datasets in long-term multivariate time-series forecasting, including ETT, Electricity, Traffic, Weather and PEMS:
\begin{itemize}
    \item The ETT~\citep{zhou2021informer} (Electricity Transformer Temperature) dataset contains two years of data from two separate countries in China with intervals of 1-hour level (ETTh) and 15-minute level (ETTm) collected from electricity transformers. Each time step contains six power load features and oil temperature.
    \item The Electricity \footnote[1]{ \href{https://archive.ics.uci.edu/ml/datasets/ElectricityLoadDiagrams20112014}{https://archive.ics.uci.edu/ml/datasets/ElectricityLoadDiagrams20112014.}} dataset describes 321 clients' hourly electricity consumption  from 2012 to 2014.
    \item The Traffic \footnote[2]{\href{http://pems.dot.ca.gov}{http://pems.dot.ca.gov.}} dataset contains the road occupancy rates from various sensors on San Francisco Bay area freeways, which is provided by California Department of Transportation.
    \item the Weather \footnote[3]{\href{https://www.bgc-jena.mpg.de/wetter/}{https://www.bgc-jena.mpg.de/wetter/.}} dataset contains 21 meteorological indicators collected at around 1,600 landmarks in the United States.
    \item the PEMS~\citep{chen2001freeway} (PEMS03, PEMS04, and PEMS08) measures the highway traffic of California in real-time every 30 seconds.
\end{itemize}
\subsection{Baselines}
{We choose SOTA and the most representative LTSF models as our baselines, including Channel-Independence models and Channel-mixing models. }
\begin{itemize}
    \item { PatchTST~\citep{nie2022time}: the current SOTA LTSF models. It utilizes channel-independence and patch techniques and achieves the highest performance by utilizing the native Transformer. We directly use the public official source code for implementation.\footnote[4]{ \href{https://github.com/yuqinie98/PatchTST} {https://github.com/yuqinie98/PatchTST}} }
    \item { Informer~\citep{zhou2021informer}: it proposes improvements to the Transformer model by utilizing the a sparse self-attention mechanism. We take the official open source code.\footnote[5]{ \href{https://github.com/zhouhaoyi/Informer2020} {https://github.com/zhouhaoyi/Informer2020}} }
    \item { NSformer~\cite{liu2022non}: NSformer is to address the over-stationary problem and it devises the De-stationary Attention to recover the intrinsic non-stationay information into temporal dependencies by approximating distinguishable attentions learned from raw series. We also take the official implemented code.\footnote[6]{ \href{https://github.com/thuml/Nonstationary_Transformers} {https://github.com/thuml/Nonstationary\_Transformers}} } 
    
    \item { Crossformer~\cite{zhang2023crossformer}: similar to PatchTST, it also utilizes the patch techniques. Unlike PatchTST, it leverages cross-variable and cross-time attention. We utilize the official code.  \footnote[7]{ \href{https://github.com/Thinklab-SJTU/Crossformer} {https://github.com/Thinklab-SJTU/Crossformer}} and fix the input length to 96.}
    \item { RMLP: it is a linear-based models which consists of two linear layers with ReLU activation.}
\end{itemize}
\label{baselines}

{For the ETT, Weather, Electricity, and Traffic datasets, we set $I=96$ for Channel-mixing models and $I=336$ for Channel-Independence models, as longer input lengths tend to yield better performance for Channel-Independence models. For PEMS03, PEMS04, and PEMS08 datasets, we set $I=336$ for all of these models since all of them perform better in a longer input length.}

\subsection{{Synthetic Data}}
\label{sec:syn}
{To demonstrate that \framework \xspace can take advantages of cross-variable correlation while avoiding unnecessary noise, we also conducted experiments on simulated data. The function for the synthetic data is $y_{i}=\sum_{j=1}^{\mathit{J}}A^{j}_{i}sin(\omega_{i}^{j}x+\varphi^{j}_{i})$ for $x_{i} \in [0,1]$,  where the frequencies, amplitude and phase shifts are randomly selected via $\omega_{i}^{j} \sim \mathcal{U}(0,\pi)$, $A_{i}^{j} \sim \mathcal{U}(0, 1)$, $\varphi_{i}^{j} \sim \mathcal{U}(0,\pi)$, and we set $\mathit{J}=5$.  Meanwhile, we add Gaussian noise $\epsilon \sim \mathcal{N}(0,\sigma)$to simulate the noise situation, which is independent of $y$. And the experimental settings are shown in Table~\ref{tab:syn_data_setting}. }


\begin{table}[h]
    \centering
    \caption{Experimental settings on Synthetic Data. }
    \begin{tabular}{|c|c|c|c|}
    \hline
         &  Channel-mixing & Channel-Independence  & \mf\\
        \hline
        \multirow{6}{*}{Input} & $i_{t}=y_{t-T:t}$ & $i_{t}=y_{t-T:t}$ & $i_{t}=y_{t-T:t}$ \\
         & Amplitude $A_{t+1:t+P}$ & &$A_{t+1:t+P}$ \\
         & Frequency $\omega_{t+1:t+P}$ & & Frequency $\omega_{t+1:t+P}$ \\
         & Phase Shifts $\varphi_{t+1:t+P}$ & & Phase Shifts $\varphi_{t+1:t+P}$ \\
         & Variable $x_{t+1:t+P}$ & & Variable $x_{t+1:t+P}$ \\
         & Noise $\epsilon$ & & Noise $\epsilon$ \\
         \hline
         Output & $o_{t}=y_{t+1:t+P}$ & $o_{t}=y_{t+1:t+P}$ & $o_{t}=y_{t+1:t+P}$ \\ 
         \hline
         Time Mixing & 3-layers MLP & 3-layers MLP & -  \\
         \hline
         Channel Mixing & 2-layers MLP & - & Directly add to CM Model  \\
         \hline
    \end{tabular}
    \label{tab:syn_data_setting}
\end{table}

\begin{figure*}
	\centering
	\subfloat[\centering Experimental results]{
		\begin{minipage}[t]{0.32\linewidth}
			\centering
			\includegraphics[width=2.0in]{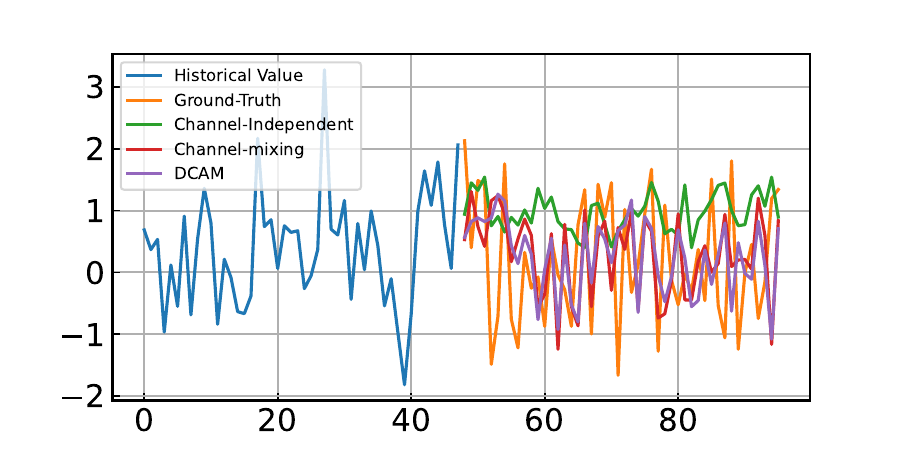}\\
			\vspace{0.00cm}
		\end{minipage}%
	}%
	\subfloat[\centering Performance under different $\sigma$ during training]{
		\begin{minipage}[t]{0.32\linewidth}
			\centering
			\includegraphics[width=2.1in]{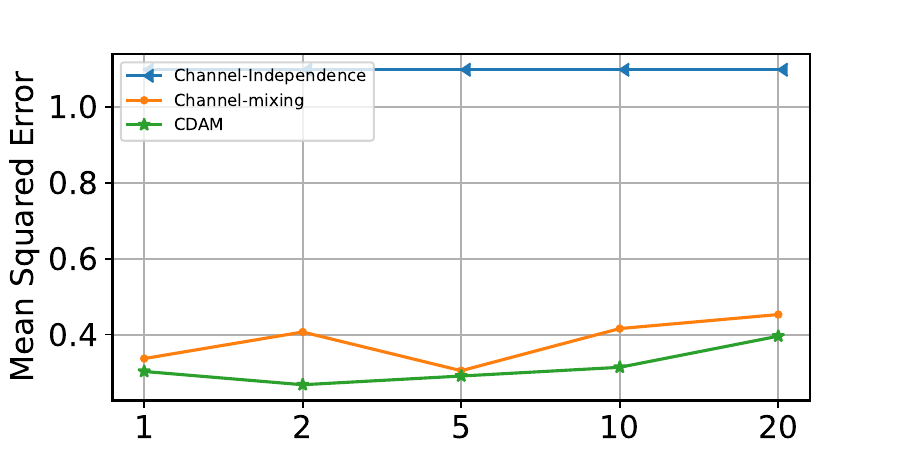}\\
			\vspace{0.00cm}
		\end{minipage}%
	}%
	\subfloat[\centering Performance under different $\sigma$ during testing]{
		\begin{minipage}[t]{0.32\linewidth}
			\centering
			\includegraphics[width=2.1in]{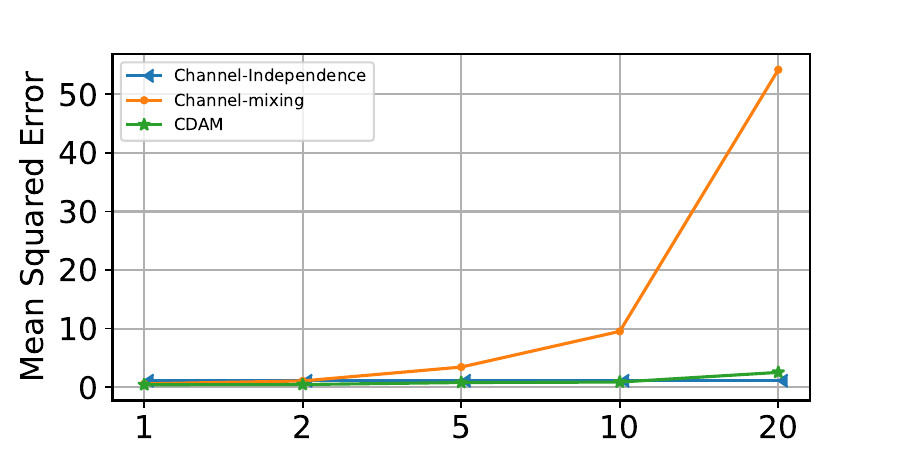}\\
			\vspace{0.00cm}
		\end{minipage}%
	}%
	\centering
	\caption{Experimental results on synthetic data.}
	\vspace{-0.2cm}
	\label{fig:syn_results}
\end{figure*}

{The experimental results on synthetic data are presented in Figure~\ref{fig:syn_results}. Since the Channel-Independence model only takes into account the historical value and variables such as $A, \omega, \varphi, x$ change over time and are not fixed, its performance is notably inferior to that of the Channel-mixing model and \mf. Additionally, we conducted experiments to investigate the impact of noise by training these three models using different noise levels, manipulated through the adjustment of $\sigma$, while maintaining a consistent $\sigma$ during testing. The experimental results are shown in Figure~\ref{fig:syn_results} (b). We observed that when the noise in the training set and the test set follows the same distribution, noise has a minimal effect on the models' performance, resulting in only a slight reduction. Notably, compared to the Channel-mixing model, our \mf \xspace model performs consistently well across various noise levels. To further demonstrate the noise reduction capability of \mf \xspace, we set $\sigma=1$ during training and modified the variance $\sigma$ during testing. As depicted in Figure~\ref{fig:syn_results} (c), we observed that the performance of \mf \xspace remains relatively stable, while the effectiveness of the Channel-mixing model is significantly impacted. This observation highlights the ability of \mf \xspace to effectively minimize the influence of irrelevant noise, despite the fact that complete elimination is not achieved. }


\subsection{\textbf{Lower bound for} $I(X^{i},Y^{i};Z^{i})$}
\subsubsection{Derivation of $I(X^{i},Y^{i};Z^{i})$}
\label{sec:a2.1}
The joint Mutual Information between the i-th historical series $X^{i}$, i-th future series $Y^{i}$ and latent representation $Z^{i}$ is defined as:
\begin{equation}
    \begin{aligned}
        I(X^{i},Y^{i};Z^{i})&=I(X^{i};Z^{i})+I(Z^{i};Y^{i}|X^{i}) \\
        &=\int p(z^{i},x^{i}) \log \frac{p(z^{i},x^{i})}{p(z^{i})p(x^{i})}dz^{i}dx^{i}+\int p(z^{i},y^{i},x^{i}) \log \frac{p(x^{i})p(z^{i},y^{i},x^{i})}{p(z^{i},x^{i})p(y^{i},x^{i})} dz^{i} dy^{i} d x^{i}  \\
        &=\int p(z^{i},y^{i},x^{i}) \log \frac{p(z^{i},x^{i})}{p(z^{i})p(x^{i})}dz^{i}dy^{i}dx^{i}+\int p(z^{i},y^{i},x^{i}) \log \frac{p(x^{i})p(z^{i},y^{i},x^{i})}{p(z^{i},x^{i})p(y^{i},x^{i})} dz^{i} dy^{i} d x^{i} \\
        &= \int p(z^{i},y^{i},x^{i}) \log \frac{p(z^{i},y^{i},x^{i})}{p(z^{i})p(y^{i},x^{i})} dz^{i} dy^{i} dx^{i} \\
        &= \int p(z^{i},y^{i},x^{i}) \log \frac{p(x^{i},y^{i}|z^{i})}{p(y^{i},x^{i})} dz^{i} dy^{i} dx^{i} \\
        &=\int p(z^{i},y^{i},x^{i}) \log p(y^{i},x^{i}|z^{i})dz^{i}dy^{i}dx^{i} -\int p(z^{i},y^{i},x^{i}) \log p(y^{i},x^{i}) dz^{i} dy^{i} dx^{i} \\
        &=\int p(z^{i},y^{i},x^{i}) \log p(y^{i}|x^{i},z^{i}) p(x^{i}|z^{i}) dz^{i} dy^{i} dx^{i} -\int p(y^{i},x^{i}) \log p(y^{i},x^{i}) dy^{i} dx^{i} \\
        &=\E_{p(z^{i},y^{i},x^{i})} \left[ \log p(y^{i}|x^{i},z^{i})\right]+\E_{p(z^{i},x^{i})} \left [ \log p(x^{i}|z^{i}) \right] + H(Y^{i},X^{i}) \\
        \end{aligned}
\end{equation}

Therefore, $I(X^{i};Y^{i};Z^{i})$ can be represented as:
\begin{equation}
    I(X^{i},Y^{i}|Z^{i})=\E_{p(z^{i},y^{i},x^{i})} \left[ \log p(y^{i}|x^{i},z^{i})\right]+\E_{p(z^{i},x^{i})} \left [ \log p(x^{i}|z^{i}) \right] + H(Y^{i},X^{i}) \\
\end{equation}

\subsubsection{Variational Approximation of $I(X^{i},Y^{i};Z^{i})$}
\label{sec:a2.2}
\begin{equation}
    \begin{aligned}
        I(X^{i},Y^{i};Z^{i})&= \E_{p(z^{i},y^{i},x^{i})} \left[ \log p(y^{i}|x^{i},z^{i})\right]+\E_{p(z^{i},x^{i})} \left [ \log p(x^{i}|z^{i}) \right]+\text{constant} \\
    \end{aligned}
\end{equation}

We introduce $p_{\theta}(x^{i}|z^{i})$ to be the variational approximation of $p(x^{i},y^{i})$. Since the Kullback Leibler (KL) divergence is always non-negetive, we have:
\begin{equation}
\begin{aligned}
    D_{KL} \left[  p(X^{i}|Z^{i}) || p_{\theta}(X^{i}|Z^{i}) \right]& =\int p(x^{i}|z^{i}) log \frac{p(x^{i}|z^{i})}{p_{\theta}(x^{i}|z^{i})}d x^{i} \geq 0 \\
    \E_{p(z^{i},x^{i})}\left[  \log p(x^{i}|z^{i}) \right] & \geq \E_{p(z^{i},x^{i})}\left[  \log p_{\theta}(x^{i}|z^{i}) \right]
\end{aligned}
\end{equation}
In the same way, we have:
\begin{equation}
    \E_{p(z^{i},y^{i},x^{i})} \left [  \log p(y^{i}|x^{i},z^{i}) \right ] \geq \E_{p(z^{i},y^{i},x^{i})} \left [  \log p_{\theta}(y^{i}|x^{i},z^{i}) \right ]
\end{equation}
Therefore, the variational lower bound is as follows:
\begin{equation}
    I(X^{i},Y^{i};Z^{i})- \text{constant} \geq I_v(X^{i},Y^{i};Z^{i})= \E_{p(z^{i},y^{i},x^{i})} \left[ \log p_{\theta}(y^{i}|x^{i},z^{i})\right]+\E_{p(z^{i},x^{i})} \left [ \log p_{\theta}(x^{i}|z^{i}) \right]
\end{equation}
 $I(X^{i},Y^{i};Z^{i})$ can thus be maximized by maximizing its variational lower bound.

\subsection{Derivation and variational approximation of $I(\hat{Y}_{j}^{i};Y_{j-1}^{i}|X^{i})$}
\label{sec:a.3}
The Mutual Information between the $j$-th downsampling predicted subsequence $\hat{Y}_{j}^{i}$ and $(j-1)$-th target downsampling subsequence $Y_{j-1}^{i}$ given historical sequence $X^{i}$ is defined as:
\begin{equation}
\begin{aligned}
    I(\hat{Y}_{j}^{i};Y_{j-1}^{i}|X^{i})&=\int p(\hat{y}^{i}_{j},y^{i}_{j-1},x^{i}) \log \frac{p(\hat{y}_{j}^{i},y_{j-1}^{i}|x^{i})}{p(\hat{y}^{i}_{j}|x^{i})p(y^{i}_{j-1}|x^{i})} d\hat{y}^{i}_{j} dy_{j-1}^{i} d x^{i} \\
\end{aligned}
\end{equation}

\begin{equation}
    \begin{aligned}
        I(Y_{j-1}^{i};\hat{Y}_{j}^{i}|X^{i})&=\int p(y_{j-1}^{i},x^{i},\hat{y}_{j}^{i}) \log \frac{p(x^{i})p(y^{i}_{j-1},\hat{y}_{j}^{i},x^{i})}{p(y_{j-1}^{i},x^{i})p(\hat{y}_{j}^{i},x^{i})} d\hat{y}_{j-1}^{i} dy_{j}^{i} dx^{i} \\
        &=\int p(y_{j-1}^{i},x^{i},\hat{y}_{j}^{i}) \log \frac{p(y_{j-1}^{i}|\hat{y}_{j}^{i},x^{i})}{p(y_{j-1}^{i}|x^{i})}d\hat{y}_{j-1}^{i} dy_{j}^{i} dx^{i} \\
        &=\int p(y_{j-1}^{i},x^{i},\hat{y}_{j}^{i}) \log p(y^{i}_{j-1}|\hat{y}^{i}_{j},x^{i}) d y_{j-1}^{i} d \hat{y}_{j}^{i} dx^{i} + H(Y_{j-1}^{i},X^{i}) \\
        & \geq \int p(y_{j-1}^{i},x^{i},\hat{y}_{j}^{i}) \log p(y^{i}_{j-1}|\hat{y}^{i}_{j},x^{i}) d y_{j-1}^{i} d\hat{y}_{j}^{i} dx^{i} \\
    \end{aligned}
\end{equation}

Since $p(y_{j-1}^{i}|\hat{y}_{j}^{i},x^{i})$ is intractable, we use $p_{\theta}(y_{j-1}^{i}|\hat{y}_{j}^{i},x^{i})$ to approximate $p(\hat{y}_{j}^{i}|y_{j-1}^{i},x^{i})$, therefore, we have:
\begin{equation}
    \begin{aligned}
        I(Y_{j-1}^{i};\hat{Y}_{j}^{i}|X^{i})& \geq \int p(y_{j-1}^{i},x^{i},\hat{y}_{j}^{i}) \log p(y^{i}_{j-1}|\hat{y}^{i}_{j},x^{i}) dy_{j-1}^{i} d\hat{y}_{j}^{i} dx^{i} \\
        & \geq \int p(y^{i}_{j-1},\hat{y}^{i}_{j},x^{i}) \log \frac{p(y^{i}_{j-1}|\hat{y}^{i}_{j},x^{i})  p_{\theta}(y^{i}_{j-1}|\hat{y}^{i}_{j},x^{i})}{p_{\theta}(y^{i}_{j-1}|\hat{y}^{i}_{j},x^{i})} dy_{j-1}^{i} d \hat{y}_{j}^{i} d x^{i} \\
        & \geq \int p(y_{j-1}^{i},\hat{y}_{j}^{i},x^{i}) \log p_{\theta} (y_{j-1}^{i}|\hat{y}_{j}^{i},x^{i}) d y_{j-1}^{i} d \hat{y}_{j}^{i} d x^{i} \\
        & \geq \E_{p(y_{j-1}^{i},\hat{y}_{j}^{i},x^{i})} \left[ p_{\theta}(y_{j-1}^{i}|\hat{y}_{j}^{i},x^{i}) \right]
    \end{aligned}
\end{equation}

Therefore, the Mutual information $I(Y_{j-1}^{i};\hat{Y}_{j}^{i}|X^{i})$ can be maximized by maximizing $\E_{p(y_{j-1}^{i},\hat{y}_{j}^{i},x^{i})} \left[ p_{\theta}(y_{j-1}^{i}|\hat{y}_{j}^{i},x^{i}) \right]$.

\subsection{Extra Experimental Results}
\subsubsection{Full Results of Channel-Independence models}
{ In this section, we provide the full experimental results of Channel-Independence models in Table~\ref{tab:main_results_channel_independent_full} which is an extended version of Table~\ref{tab:marin_results_channel_independent}}
\begin{table*}[h]
\caption{Multivariate long-term series forecasting results on Channel-Independence models with different prediction lengths $O \in \{96,192,336,720\}$. We set the input length $I$ as 336 for all the models. The best result is indicated in bold font. ({ \textit{Avg} is averaged from all four prediction lengths and \textit{Pro} means the relative MSE and MAE reduction.})}
    \centering
    \begin{tabular}{c|c|cccc|cccc}
    \hline
    \multicolumn{2}{c|}{\multirow{2}{*}{Models}} & \multicolumn{4}{c|}{PatchTST} & \multicolumn{4}{c}{RMLP}  \\
    
    \multicolumn{2}{c|}{} & \multicolumn{2}{c}{Original} & \multicolumn{2}{c|}{w/Ours} & \multicolumn{2}{c}{Original} & \multicolumn{2}{c}{w/Ours}  \\
    \hline
    \multicolumn{2}{c|}{Metric} & MSE & MAE & MSE & MAE & MSE & MAE & MSE & MAE  \\
    \hline
    \multirow{4}{*}{\rotatebox{90}{ETTh1}} 
    & 96 & 0.375 & 0.399 & \textbf{0.365} & \textbf{0.389} & 0.380 & 0.401 & \textbf{0.367} & \textbf{0.391} \\
    & 192 & 0.414 & 0.421 & \textbf{0.403} & \textbf{0.413} & 0.414 & 0.421 & \textbf{0.404} & \textbf{0.413}  \\
    & 336 & 0.440 & 0.440 & \textbf{0.427} & \textbf{0.428} & 0.439 & 0.436 & \textbf{0.426} & \textbf{0.429}  \\
    & 720 & 0.460 & 0.473 & \textbf{0.433} & \textbf{0.453} & 0.470 & 0.471 & \textbf{0.439} & \textbf{0.459}  \\
    \hline
    & Avg & 0.422 & 0.433 & \textbf{0.407} & \textbf{0.420} & 0.426 & 0.432 & \textbf{0.409} & \textbf{0.423}  \\
    \hline
    & Pro & - & - & \textbf{3.5\%} & \textbf{3.0\%} & - & - & \textbf{3.9\%} & \textbf{2.1\%}  \\
    \hline
    \multirow{4}{*}{\rotatebox{90}{{ETTh2}}} 
    & 96 & 0.274 & 0.335 & \textbf{0.271} & \textbf{0.332} & 0.290 & 0.348 & \textbf{0.271} & \textbf{0.333} \\
    & 192 & 0.342 & 0.382 & \textbf{0.334} & \textbf{0.373} & 0.350 & 0.388 & \textbf{0.335} & \textbf{0.374}  \\
    & 336 & 0.365 & 0.404 & \textbf{0.357} & \textbf{0.395} & 0.374 & 0.410 & \textbf{0.358} & \textbf{0.395}  \\
    & 720 & 0.391 & 0.428 & \textbf{0.385} & \textbf{0.421} & 0.410 & 0.439 & \textbf{0.398} & \textbf{0.432}  \\
    \hline
    & Avg & 0.343 & 0.387 & \textbf{0.337} & \textbf{0.380} & 0.356 & 0.396 & \textbf{0.34} & \textbf{0.384}  \\
    \hline
    & Pro & - & - & \textbf{1.7\%} & \textbf{1.8\%} & - & - & \textbf{4.5\%} & \textbf{3.0\%}  \\
    \hline
    \multirow{4}{*}{\rotatebox{90}{ETTm1}} 
    & 96 & 0.290 & 0.342 & \textbf{0.283} & \textbf{0.335} & 0.290 & 0.343 & \textbf{0.285} & \textbf{0.335} \\
    & 192 & 0.332 & 0.369 & \textbf{0.322} & \textbf{0.359} & 0.329 & 0.368 & \textbf{0.322} & \textbf{0.359}  \\
    & 336 & 0.366 & 0.392 & \textbf{0.356} & \textbf{0.382} & 0.364 & 0.390 & \textbf{0.358} & \textbf{0.381}  \\
    & 720 & 0.420 & 0.424 & \textbf{0.407} & \textbf{0.417} & 0.430 & 0.426 & \textbf{0.414} & \textbf{0.413} \\
    \hline
    & Avg & 0.352 & 0.381 & \textbf{0.342} & \textbf{0.373} & 0.353 & 0.381 & \textbf{0.344} & \textbf{0.372}  \\
    \hline
    & Pro & - & - & \textbf{2.8\%} & \textbf{2.1\%} & - & - & \textbf{2.5\%} & \textbf{2.3\%}  \\
    \hline
    \multirow{4}{*}{\rotatebox{90}{ETTm2}}
    & 96 & 0.165 & 0.255 & \textbf{0.161} & \textbf{0.250} & 0.177 & 0.263 & \textbf{0.162} & \textbf{0.252} \\
    & 192 & 0.220 & 0.292 & \textbf{0.217} & \textbf{0.289} & 0.233 & 0.302 & \textbf{0.217} & \textbf{0.289} \\
    & 336 & 0.278 & 0.329 & \textbf{0.271} & \textbf{0.324} & 0.283 & 0.335 & \textbf{0.270} & \textbf{0.324} \\
    & 720 & 0.367 & 0.385 & \textbf{0.362} & \textbf{0.381} & 0.367 & 0.388 & \textbf{0.357} & \textbf{0.380} \\
    \hline
    & Avg & 0.257 & 0.315 & \textbf{0.252} & \textbf{0.311} & 0.265 & 0.322 & \textbf{0.251} & \textbf{0.311}  \\
    \hline
    & Pro & - & - & \textbf{1.9\%} & \textbf{1.3\%} & - & - & \textbf{5.2\%} & \textbf{3.4\%}  \\
    \hline
    \multirow{4}{*}{\rotatebox{90}{Weather}} 
    & 96 & 0.152 & 0.199 & \textbf{0.144} & \textbf{0.194} & 0.147 & 0.198 & \textbf{0.144} & \textbf{0.196}  \\
    & 192 & 0.197 & 0.243 & \textbf{0.189} & \textbf{0.238} & 0.190 & 0.239 & \textbf{0.187} & \textbf{0.237} \\
    & 336 & 0.250 & 0.284 & \textbf{0.239} & \textbf{0.279} & 0.243 & 0.280 & \textbf{0.239} & \textbf{0.277} \\
    & 720 & 0.320 & 0.335 & \textbf{0.312} & \textbf{0.331} & 0.320 & 0.332 & \textbf{0.316} & \textbf{0.330}  \\
    \hline
    & Avg & 0.229 & 0.265 & \textbf{0.221} & \textbf{0.260} & 0.225 & 0.262 & \textbf{0.221} & \textbf{0.260}  \\
    \hline
    & Pro & - & - & \textbf{3.5\%} & \textbf{1.8\%} & - & - & \textbf{1.6\%} & \textbf{0.9\%}  \\
    \hline
    \multirow{4}{*}{\rotatebox{90}{Traffic}}
    & 96 & 0.367 & 0.251 & \textbf{0.358} & \textbf{0.245} & 0.383 & 0.269 & \textbf{0.364} & \textbf{0.249} \\
    & 192 & 0.385 & 0.259 & \textbf{0.379} & \textbf{0.254} & 0.401 & 0.276 & \textbf{0.384} & \textbf{0.258}  \\
    & 336 & 0.398 & 0.265 & \textbf{0.391} & \textbf{0.261} & 0.414 & 0.282 & \textbf{0.398} & \textbf{0.266}  \\
    & 720 & 0.434 & 0.287 & \textbf{0.425} & \textbf{0.280} & 0.443 & 0.309 & \textbf{0.428} & \textbf{0.284} \\
    \hline
    & Avg & 0.396 & 0.265 & \textbf{0.388} & \textbf{0.260} & 0.410 & 0.284 & \textbf{0.393} & \textbf{0.264}  \\
    \hline
    & Pro & - & - & \textbf{2.0\%} & \textbf{1.8\%} & - & - & \textbf{5.2\%} & \textbf{8.4\%}  \\
    \hline
    \multirow{4}{*}{\rotatebox{90}{Electricity}} & 96 & 0.130 & 0.222 & \textbf{0.125} & \textbf{0.219} & 0.130 & 0.225 & \textbf{0.125} & \textbf{0.218} \\
    & 192 & 0.148 & 0.242 & \textbf{0.143} & \textbf{0.235} & 0.148 & 0.240 & \textbf{0.144} & \textbf{0.236}  \\
    & 336 & 0.167 & 0.261 & \textbf{0.161} & \textbf{0.255} & 0.164 & 0.257 & \textbf{0.160} & \textbf{0.253}  \\
    & 720 & 0.202 & 0.291 & \textbf{0.198} & \textbf{0.287} & 0.203 & 0.291 & \textbf{0.195} & \textbf{0.285}  \\
    \hline
    & Avg & 0.161 & 0.254 & \textbf{0.156} & \textbf{0.249} & 0.161 & 0.253 & \textbf{0.156} & \textbf{0.248}  \\
    \hline
    & Pro & - & - & \textbf{3.1\%} & \textbf{1.9\%} & - & - & \textbf{3.1\%} & \textbf{2.1\%}  \\
    \hline
    \end{tabular}
    \label{tab:main_results_channel_independent_full}
\end{table*}

\subsubsection{Additional Ablation Study}
{In this section, we provide the full ablation experimental results of the Channel-mixing models and Channel-Independence models in Table~\ref{tab:ablation_dependent} and Table~\ref{tab:ablation_independent}, respectively, which are the extended of Table~\ref{tab:ablation}. We also provide Table~\ref{tab:pems_ablation_results}, which contains the full results of ablation experiments on the PEMS (PEMS03, PEMS04, PEMS08) dataset.}
\begin{table*}[h]\normalsize
\centering
\tabcolsep=0.15cm
\caption{{Component ablation of \framework \xspace for RMLP and PatchTST. We set the input length $I$ as 336.  The best results are in \textbf{bold} and the second best are \underline{underlined}.}}
\begin{tabular}{cc|cccccc|cccccc}
\hline
     \multicolumn{2}{c|}{\multirow{2}{*}{Models}} & \multicolumn{6}{c|}{RMLP} & \multicolumn{6}{c}{PatchTST} \\
     \multicolumn{2}{c|}{} & \multicolumn{2}{c}{Original} & \multicolumn{2}{c}{+\ms} & \multicolumn{2}{c|}{+\framework} & \multicolumn{2}{c}{Original} & \multicolumn{2}{c}{+\ms} & \multicolumn{2}{c}{+\framework} \\
     \hline
     \multicolumn{2}{c|}{Metric} & MSE & MAE & MSE & MAE & MSE & MAE & MSE & MAE & MSE & MAE & MSE & MAE \\
     \hline 
     \multirow{4}{*}{\rotatebox{90}{ETTh1}} 
     & 96 & 0.380 & 0.401 & \underline{0.371} & \underline{0.392} & \textbf{0.367} & \textbf{0.391} & 0.375 & 0.399 & \underline{0.367} & \underline{0.391} & \textbf{0.365} & \textbf{0.389} \\ 
    &192 & 0.414 & 0.421 & \underline{0.406} & \underline{0.414} & \textbf{0.404} & \textbf{0.413} & 0.414 & 0.421 & \underline{0.405} & \underline{0.414} & \textbf{0.403}& \textbf{0.413}\\ 
    & 336 & 0.439 & 0.436 & \underline{0.427} & \textbf{0.428} & \textbf{0.426} & \underline{0.429} & 0.440 & 0.440 & \underline{0.429} & \underline{0.430} & \textbf{0.427} & \textbf{0.428} \\ 
    & 720 & 0.470 & 0.471 & \underline{0.450} & \underline{0.465} & \textbf{0.439} & \textbf{0.459 }& 0.460 & 0.473 & \underline{0.435} & \underline{0.455} & \textbf{0.433} & \textbf{0.453}\\
    \hline 
    \multirow{4}{*}{\rotatebox{90}{{ETTh2}}} 
     & 96 & 0.290 & 0.348 & \underline{0.278} & \underline{0.337} & \textbf{0.271} & \textbf{0.333} & 0.274 & 0.335 & \textbf{0.271} & \textbf{0.332} & \textbf{0.271} & \textbf{0.332} \\ 
    & 192 & 0.350 & 0.388 & \underline{0.340} & \underline{0.377} & \textbf{0.335} & \textbf{0.374} & 0.342 & 0.382 & \textbf{0.334} & \textbf{0.373} & \textbf{0.334}& \textbf{0.373}\\ 
    & 336 & 0.374 & 0.410 & \underline{0.366} & \underline{0.402} & \textbf{0.358} & \underline{0.395} & 0.365 & 0.404 & \textbf{0.357} & \textbf{0.393} & \textbf{0.357} & \underline{0.395} \\ 
    & 720 & 0.410 & 0.439 & \underline{0.404} & \underline{0.435} & \textbf{0.398} & \textbf{0.432 }& 0.391 & 0.428 & \underline{0.386} & \underline{0.422} & \textbf{0.385} & \textbf{0.421}\\
    \hline 
     \multirow{4}{*}{\rotatebox{90}{ETTm1}} 
     & 96  & 0.290 & 0.343 & \textbf{0.285} & \underline{0.337} & \textbf{0.285} & \textbf{0.335} & 0.290 & 0.342 & \underline{0.286} & \underline{0.337} & \textbf{0.283} & \textbf{0.335} \\ 
       &  192 & 0.329 & 0.368 & \textbf{0.321} & \underline{0.360} & \underline{0.322} & \textbf{0.359} & 0.332 & 0.369 & \underline{0.326} & \underline{0.366} & \textbf{0.322} & \textbf{0.359} \\ 
       &  336 & 0.364 & 0.390 & \textbf{0.357} & \underline{0.382} & \underline{0.358} & \textbf{0.381} & 0.366 & 0.392 & \underline{0.359} & \underline{0.388} & \textbf{0.356}& \textbf{0.382} \\ 
        & 720 & 0.430 & 0.426 & \underline{0.415} & \underline{0.415} & \textbf{0.414} & \textbf{0.413} & 0.420 & 0.424 & \underline{0.408} & \textbf{0.413} & \textbf{0.407} & \underline{0.417} \\ 
      
    \hline 
     \multirow{4}{*}{\rotatebox{90}{ETTm2}} 
     & 96 & 0.177 & 0.263 & \underline{0.166} & \underline{0.255} & \textbf{0.162} & \textbf{0.252} & 0.165 & 0.255 & \underline{0.162} & \underline{0.251} & \textbf{0.161} & \textbf{0.250} \\
       & 192 & 0.233 & 0.302 & \underline{0.222} & \underline{0.294} & \textbf{0.217} & \textbf{0.289} & 0.220 & 0.292 & \underline{0.218} & \underline{0.290} & \textbf{0.217} &\textbf{0.289} \\ 
       & 336 & 0.283 & 0.335 & \underline{0.274} & \underline{0.328} &\textbf{0.270} & \textbf{0.324} & 0.278 & 0.329 & \underline{0.273} & \underline{0.326} & \textbf{0.271} & \textbf{0.324} \\ 
       & 720 & 0.367 & 0.388 & \underline{0.362} & \underline{0.384} & \textbf{0.357} & \textbf{0.380} & 0.367 & 0.385 & \underline{0.364} & \underline{0.382} & \textbf{0.362} & \textbf{0.381} \\
     \hline 
     \multirow{4}{*}{\rotatebox{90}{Weather}} 
     & 96 & 0.147 & 0.198 & \underline{0.146} & \underline{0.197} & \textbf{0.144} & \textbf{0.196} & 0.152 & 0.199 & \underline{0.149} & \underline{0.197} & \textbf{0.144} & \textbf{0.194} \\ 
        &192 & 0.190 & 0.239 & \underline{0.189} & \underline{0.238} & \textbf{0.187}& \textbf{0.237 }& 0.197 & 0.243 & \underline{0.192} & \textbf{0.238} & \textbf{0.189} & \textbf{0.238} \\ 
       & 336 & 0.243 & 0.280 & \underline{0.241} & \underline{0.278} & \textbf{0.239} & \textbf{0.277} & 0.250 & 0.284 & \underline{0.247} & \underline{0.280} & \textbf{0.239} & \textbf{0.279} \\
       & 720 & \underline{0.320} & 0.332 & \underline{0.319} & \underline{0.332} & \textbf{0.316} & \textbf{0.330} & 0.320 & 0.335 & 0.321 & \underline{0.332} & \textbf{0.312}& \textbf{0.331} \\
     \hline 
     \multirow{4}{*}{\rotatebox{90}{Traffic}} 
     & 96 & 0.383 & 0.269 & \underline{0.366} & \underline{0.252} & \textbf{0.364} & \textbf{0.249} & 0.367 & 0.251 & \underline{0.359} & \textbf{0.245} & \textbf{0.358} & \textbf{0.245 }\\ 
      &  192 & 0.401 & 0.276 & \underline{0.386} & \underline{0.260} & \textbf{0.384} & \textbf{0.258} & 0.385 & 0.259 & \underline{0.380} & \underline{0.255} &\textbf{0.379} & \textbf{0.254} \\ 
      &  336 & 0.414 & 0.282 & \underline{0.400} & \underline{0.268} & \textbf{0.398} & \textbf{0.266} & 0.398 & 0.265 & \textbf{0.391} & \underline{0.262} & \textbf{0.391} & \textbf{0.261} \\ 
       & 720 & 0.443 & 0.309 & \underline{0.432} & \underline{0.286} &\textbf{0.428} & \textbf{0.284} & 0.434 & 0.287 & \textbf{0.424} & \textbf{0.279} & \underline{0.425} & \underline{0.280} \\ 
     \hline 
     \multirow{4}{*}{\rotatebox{90}{Electricity}} 
     & 96 & 0.130 & 0.225 & 0.127 & 0.221 & \textbf{0.125} & \textbf{0.218} & 0.130 & \underline{0.222} & \underline{0.129} & 0.223 &\textbf{0.125} & \textbf{0.219} \\ 
       & 192 & 0.148 & 0.240 & 0.145 & 0.238 & \textbf{0.144} & \textbf{0.236} & 0.148 & 0.242 & \underline{0.147} & \underline{0.240} & \textbf{0.143} & \textbf{0.235} \\ 
       & 336 & 0.164 & 0.257 & 0.162 & 0.255 & \textbf{0.160} & \textbf{0.253} & 0.167 & 0.261 & \underline{0.165} & \underline{0.258} & \textbf{0.161} & \textbf{0.255} \\ 
      &  720 & 0.203 & 0.291 & 0.199 & 0.288 & \textbf{0.195} & \textbf{0.285} & \underline{0.202} & \underline{0.291} & 0.204 & 0.292 & \textbf{0.198} & \textbf{0.287} \\ \hline
      \label{tab:ablation_independent}
\end{tabular}
\end{table*}

\begin{table*}[h]\normalsize
\centering
\tabcolsep=0.15cm
\caption{{Component ablation of \framework \xspace for Informer, Stationary, and Crossformer. We set the input length $I$ as 96.  The best results are in \textbf{bold} and the second best are \underline{underlined}.}}
\resizebox{1\linewidth}{!}{
\begin{tabular}{cc|cccccc|cccccc|cccccc}
\hline
     \multicolumn{2}{c|}{\multirow{2}{*}{Models}} & \multicolumn{6}{c|}{Informer} & \multicolumn{6}{c|}{Stationary} & \multicolumn{6}{c}{Crossformer} \\
     \multicolumn{2}{c|}{} & \multicolumn{2}{c}{Original} & \multicolumn{2}{c}{+\ms} & \multicolumn{2}{c|}{+\framework} & \multicolumn{2}{c}{Original} & \multicolumn{2}{c}{+\ms} & \multicolumn{2}{c|}{+\framework} & \multicolumn{2}{c}{Original} & \multicolumn{2}{c}{+\ms} & \multicolumn{2}{c}{+\framework} \\
     \hline
     \multicolumn{2}{c|}{Metric} & MSE & MAE & MSE & MAE & MSE & MAE & MSE & MAE & MSE & MAE & MSE & MAE  & MSE & MAE & MSE & MAE & MSE & MAE \\
      \hline 
     \multirow{4}{*}{\rotatebox{90}{ETTh1}} 
     & 96 & 0.865 & 0.713 & \underline{0.598} & \underline{0.565} & \textbf{0.381} & \textbf{0.394}& 0.598 & 0.498 & \underline{0.455} & \underline{0.452} & \textbf{0.375} & \textbf{0.388} & 0.457 & 0.463 & \underline{0.396} & \underline{0.411} & \textbf{0.379 }& \textbf{0.392} \\ 
       & 192 & 1.008 & 0.792 & \underline{0.694} & \underline{0.640} & \textbf{0.435} & \textbf{0.430} & 0.602 & 0.520 & \underline{0.491} & \underline{0.478} & \textbf{0.425}& \textbf{0.417} & 0.635 & 0.581 & \underline{0.541} & \underline{0.511} & \textbf{0.433} & \textbf{0.427} \\ 
       & 336 & 1.107 & 0.809 & \underline{0.853} & \underline{0.719} & \textbf{0.485} & \textbf{0.461} & 0.677 & 0.573 & \underline{0.611} & \underline{0.530} & \textbf{0.463}& \textbf{0.436} & 0.776 & 0.667 & \underline{0.759} & \underline{0.651} & \textbf{0.482} & \textbf{0.458} \\ 
      & 720 & 1.181 & 0.865 & \underline{0.914} & \underline{0.741} & \textbf{0.534} & \textbf{0.524} & 0.719 & 0.597 & \underline{0.594} & \underline{0.542} & \textbf{0.463} & \textbf{0.459}& 0.861 & 0.725 & \underline{0.845} & \underline{0.711} & \textbf{0.529} & \textbf{0.517}\\
      \hline
    \multirow{4}{*}{\rotatebox{90}{{ETTh2}}}
    &96 &  3.755 & 1.525 & \underline{0.502} & \underline{0.538} & \textbf{0.336} & \textbf{0.390} & 0.362 & 0.393 & \underline{0.330} & \underline{0.371} & \textbf{0.286} & \textbf{0.335} & 0.728 & 0.615 & \underline{0.364} & \underline{0.415} & \textbf{0.333} & \textbf{0.386} \\
     & 192 &  5.602 & 1.931 & \underline{0.821} & \underline{0.701} & \textbf{0.468} & \textbf{0.470} & 0.481 & 0.453 & \underline{0.456} & \underline{0.440} & \textbf{0.371} & \textbf{0.388} & 0.898 & 0.705 & \underline{0.470} & \underline{0.481} & \textbf{0.455} & \textbf{0.453} \\ 
    &  336 &  4.721 & 1.835 & \underline{1.065} & \underline{0.823} & \textbf{0.582} & \textbf{0.534} & 0.524 & 0.487 & \underline{0.475} & \underline{0.463} &\textbf{0.414} & \textbf{0.425} & 1.132 & 0.807 & \underline{0.580} & \underline{0.547} & \textbf{0.554} & \textbf{0.513} \\ 
    & 720 &  3.647 & 1.625 & \underline{1.489} & \underline{1.022} & \textbf{0.749} & \textbf{0.620} & 0.512 & 0.494 & \underline{0.506} & \underline{0.486} & \textbf{0.418} & \textbf{0.437} & 4.390 & 1.795 & \underline{0.768}& \underline{0.648} & \textbf{0.757} & \textbf{0.619}   \\ \hline
    
     \multirow{4}{*}{\rotatebox{90}{ETTm1}} 
     &96 & 0.672 & 0.571 & \underline{0.435} & \underline{0.444} & \textbf{0.326} & \textbf{0.367} & 0.396 & 0.401 & \underline{0.375} & \underline{0.396} & \textbf{0.326} & \textbf{0.362} & 0.385 & 0.409 & \underline{0.388} & \underline{0.401} & \textbf{0.323 }& \textbf{0.362 }\\ 
       & 192 & 0.795 & 0.669 & \underline{0.473} & \underline{0.467} & \textbf{0.371}& \textbf{0.391} & 0.471 & 0.436 & \underline{0.441} & \underline{0.432} & \textbf{0.366} & \textbf{0.379} & 0.459 & 0.478 & \underline{0.436} & \underline{0.428} & \textbf{0.366} & \textbf{0.386} \\ 
       & 336 & 1.212 & 0.871 & \underline{0.545} & \underline{0.518} & \textbf{0.408} & \textbf{0.416} & 0.517 & 0.464 & \underline{0.472} & \underline{0.455} & \textbf{0.392} & \textbf{0.398} & 0.645 & 0.583 & \underline{0.483} & \underline{0.457} & \textbf{0.403} & \textbf{0.414} \\ 
       & 720 & 1.166 & 0.823 & \underline{0.669} & \underline{0.589} &\textbf{0.482} & \textbf{0.464} & 0.664 & 0.527 & \underline{0.532} & \underline{0.489} & \textbf{0.455} & \textbf{0.434} & 0.756 & 0.669 & \underline{0.548} & \underline{0.498} & \textbf{0.473 }& \textbf{0.460} \\ 
    \hline 
     \multirow{4}{*}{\rotatebox{90}{ETTm2}} 
     &96 & 0.365 & 0.453 & \underline{0.258} & \underline{0.378} & \textbf{0.187} & \textbf{0.282} & 0.201 & 0.291 & \underline{0.185} & \underline{0.276} & \textbf{0.175} & \textbf{0.256} & 0.281 & 0.373 & \underline{0.223} & \underline{0.321} & \textbf{0.186}& \textbf{0.281} \\ 
        &192 & 0.533 & 0.563 & \underline{0.439} & \underline{0.515} & \textbf{0.277} & \textbf{0.351} & 0.275 & 0.335 & \underline{0.254} & \underline{0.318} & \textbf{0.238} & \textbf{0.297} & 0.549 & 0.520 & \underline{0.347} & \underline{0.404} &\textbf{0.269} & \textbf{0.341} \\ 
       & 336 & 1.363 & 0.887 & \underline{0.836} & \underline{0.728} & \textbf{0.380} & \textbf{0.420} & 0.350 & 0.377 & \underline{0.343} & \underline{0.372} & \textbf{0.299} & \textbf{0.336} & 0.729 & 0.603 & \underline{0.528} & \underline{0.506} & \textbf{0.356} & \textbf{0.396} \\ 
       & 720 & 3.379 & 1.338 & \underline{3.172} & \underline{1.322} & \textbf{0.607} & \textbf{0.549} & 0.460 & 0.435 & \underline{0.440} & \underline{0.421} & \textbf{0.398} & \textbf{0.393} & 1.059 & 0.741 & \underline{0.895} & \underline{0.665} & \textbf{0.493} & \textbf{0.482} \\ 
     \hline 
     \multirow{4}{*}{\rotatebox{90}{Weather}} 
     & 96 & 0.300 & 0.384 & \underline{0.277} & \underline{0.354} & \textbf{0.179} & \textbf{0.249} & 0.181 & 0.230 & \underline{0.178} & \underline{0.226} & \textbf{0.166} & \textbf{0.213} & 0.158 & 0.236 & \underline{0.154} & \underline{0.225} & \textbf{0.149} & \textbf{0.218} \\ 
    & 192 & 0.598 & 0.544 & \underline{0.407} & \underline{0.447} & \textbf{0.226} & \textbf{0.296} & 0.286 & 0.312 & \underline{0.261} & \underline{0.296} & \textbf{0.218} & \textbf{0.260} & 0.209 & 0.285 & \textbf{0.202} & \textbf{0.27} & \textbf{0.202} & \underline{0.272} \\ 
     & 336 & 0.578 & 0.523 & \underline{0.529} & \underline{0.520} & \textbf{0.276} & \textbf{0.334} & 0.319 & 0.335 & \underline{0.318} & \underline{0.333} & \textbf{0.274} & \textbf{0.300} & 0.265 & 0.328 & \underline{0.263} & \underline{0.320} & \textbf{0.256}& \textbf{0.313} \\ 
     &  720 & 1.059 & 0.741 & \underline{0.951} & \underline{0.734} & \textbf{0.332} & \textbf{0.372} & 0.411 & 0.393 & \underline{0.387} & \underline{0.378} & \textbf{0.351} & \textbf{0.353} & 0.376 & 0.397 & \underline{0.353} & \underline{0.382} & \textbf{0.329} & \textbf{0.366} \\ 
     \hline 
     \multirow{4}{*}{\rotatebox{90}{Traffic}} 
     & 96 & 0.719 & 0.391 & \underline{0.577} & \underline{0.356} & \textbf{0.505} & \textbf{0.348} & 0.599 & 0.332 & \underline{0.503} & \underline{0.313} & \textbf{0.459}& \textbf{0.311} & 0.609 & 0.362 & \textbf{0.490} & \textbf{0.308} & \underline{0.529 }& \underline{0.334} \\ 
    &192 & 0.696 & 0.379 & \underline{0.556} & \underline{0.357} & \textbf{0.521} & \textbf{0.354} & 0.619 & 0.341 & \underline{0.488} & \underline{0.309} & \textbf{0.475} & \textbf{0.315} & 0.623 & 0.365 & \textbf{0.493} & \textbf{0.310} & \underline{0.519} & \underline{0.327} \\ 
    &336 & 0.777 & 0.420 & \underline{0.580} & \underline{0.370} & \textbf{0.520} & \textbf{0.337} & 0.651 & 0.347 & \underline{0.506} & \underline{0.318} & \textbf{0.486} & \textbf{0.319} & 0.649 & 0.370 & \underline{0.53} & \underline{0.328} & \textbf{0.521} & \textbf{0.337} \\ 
    & 720 & 0.864 & 0.472 & \underline{0.668} & \underline{0.430} & \textbf{0.552} & \textbf{0.352} & 0.658 & 0.358 & \underline{0.542} & \underline{0.329} & \textbf{0.522} & \textbf{0.338}& 0.758 & 0.418 & \underline{0.591} & \underline{0.348} & \textbf{0.556} & \textbf{0.350} \\ 
     \hline 
     \multirow{4}{*}{\rotatebox{90}{Electricity}} 
     &96 & 0.274 & 0.368 & \underline{0.228} & \underline{0.333} & \textbf{0.195} & \textbf{0.300} & 0.168 & 0.271 & \textbf{0.152} & \textbf{0.252} & \underline{0.154} & \underline{0.256} & 0.170 & 0.279 & \underline{0.151} & \underline{0.251} & \textbf{0.150} & \textbf{0.248} \\
    & 192 & 0.296 & 0.386 & \underline{0.238} & \underline{0.344} & \textbf{0.193} & \textbf{0.291} & 0.186 & 0.285 & \underline{0.166} & \underline{0.265} & \textbf{0.163} & \textbf{0.263} & 0.198 & 0.303 & \textbf{0.168} & \underline{0.266} & \textbf{0.168} & \textbf{0.263} \\ 
    &336 & 0.300 & 0.394 & \underline{0.254} & \underline{0.358} & \textbf{0.206} & \textbf{0.300} & 0.194 & 0.297 & \underline{0.180} & \underline{0.280} & \textbf{0.178} & \textbf{0.279} & 0.235 & 0.328 & \textbf{0.197} & \underline{0.292} & \underline{0.200} & \textbf{0.290} \\ 
    &720 & 0.373 & 0.439 & \underline{0.288} & \underline{0.379} & \textbf{0.241} & \textbf{0.332} & 0.224 & 0.316 & \underline{0.208} & \underline{0.305} & \textbf{0.201}& \textbf{0.299} & 0.27 & 0.36 & \underline{0.238} & \underline{0.328} & \textbf{0.235} & \textbf{0.323} \\ \hline 
      \label{tab:ablation_dependent}
\end{tabular}}
\end{table*}

\begin{table}[h]
    \centering
    \caption{{Component ablation of InfoTime for PatchTST, RMLP, Informer, Stationary, and Crossformer on PEMS (PEMS03, PEMS04, and PEMS08) datasets. We set the input length $I$ as 336 for all of these base models.} }
    \tabcolsep=0.08cm
    \begin{tabular}{ccc|cccc|cccc|cccc}
    \hline
     \multicolumn{2}{c}{\multirow{2}{*}{Models}} & \multirow{2}{*}{Metric} & \multicolumn{4}{c|}{PEMS03} & \multicolumn{4}{c|}{PEMS04} & \multicolumn{4}{c}{PEMS08} \\
     \multicolumn{2}{c}{} & & 96 & 192 & 336 & 720 & 96 & 192 & 336 & 720 & 96 & 192 & 336 & 720 \\
     \hline
     \multirow{6}{*}{\rotatebox{90}{PatchTST}} & \multirow{2}{*}{Original} & MSE & 0.180 & 0.207 & 0.223 & 0.291 & 0.195 & 0.218 & 0.237 & 0.321 & 0.239 & 0.292 & 0.314 & 0.372 \\
          & & MAE &  0.281 & 0.295 & 0.309 & 0.364 & 0.296 & 0.314 & 0.329 & 0.394 & 0.324 & 0.351 & 0.374 & 0.425 \\
    & \multirow{2}{*}{\textbf{+\ms}} & MSE & 0.159 & 0.189 & 0.193 & 0.263 & 0.170 & 0.198 & 0.204 & 0.257 & 0.186 & 0.244 & 0.257 & 0.307 \\
          & & MAE &  0.270 & 0.293 & 0.286 & 0.350 & 0.276 & 0.297 & 0.299 & 0.345 & 0.289 & 0.324 & 0.320 & 0.378 \\
          & \multirow{2}{*}{\textbf{+\framework}} & MSE & 
          \textbf{0.115} & \textbf{0.154} & \textbf{0.164} & \textbf{0.198} & \textbf{0.110} & \textbf{0.118} & \textbf{0.129} & \textbf{0.149} & \textbf{0.114} & \textbf{0.160} & \textbf{0.177} & \textbf{0.209}\\
          & & MAE & \textbf{0.223} & \textbf{0.251} & \textbf{0.256} & \textbf{0.286} & \textbf{0.221} & \textbf{0.224} & \textbf{0.237} & \textbf{0.261} & \textbf{0.218} & \textbf{0.243} & \textbf{0.241} & \textbf{0.281} \\ 
          \hline
          \multirow{6}{*}{\rotatebox{90}{RMLP}} & \multirow{2}{*}{Original} & MSE & 0.160 & 0.184 & 0.201 & 0.254 & 0.175 & 0.199 & 0.210 & 0.255 & 0.194 & 0.251 & 0.274 & 0.306 \\
          & & MAE  & 0.257 & 0.277 & 0.291 & 0.337 & 0.278 & 0.294 & 0.306 & 0.348 & 0.279 & 0.311 & 0.328 & 0.365 \\
          & \multirow{2}{*}{\textbf{+\ms}} & MSE & 0.143 & 0.171 & 0.186 & 0.234 & 0.153 & 0.181 & 0.189 & 0.222 & 0.158 & 0.215 & 0.236 & 0.264 \\
          & & MAE &  0.241 & 0.264 & 0.276 & 0.316 & 0.259 & 0.280 & 0.289 & 0.321 & 0.255 & 0.288 & 0.302 & 0.333 \\
          & \multirow{2}{*}{\textbf{+\framework}} & MSE & \textbf{0.117} & \textbf{0.159} & \textbf{0.146} & \textbf{0.204} & \textbf{0.103} & \textbf{0.114} & \textbf{0.130} & \textbf{0.154} & \textbf{0.116} & \textbf{0.156} & \textbf{0.175} & \textbf{0.181} \\
          & & MAE & \textbf{0.228} & \textbf{0.252} & \textbf{0.246} & \textbf{0.285} & \textbf{0.211} & \textbf{0.219} & \textbf{0.236} & \textbf{0.264} & \textbf{0.215} & \textbf{0.235} & \textbf{0.242} & \textbf{0.255} \\
          \hline
         \multirow{6}{*}{\rotatebox{90}{Informer}} & \multirow{2}{*}{Original} & MSE & 0.139 & 0.152 & 0.165 & 0.216 & 0.132 & 0.146 & 0.147 & 0.145 & 0.156 & 0.175 & 0.187 & 0.264 \\
          & & MAE  & 0.240 & 0.252 & 0.260 & 0.290 & 0.238 & 0.249 & 0.247 & 0.245 & 0.262 & 0.266 & 0.274 & 0.325 \\
          & \multirow{2}{*}{\textbf{+\ms}} & MSE & 0.126 & 0.142 & 0.157 & 0.207 & 0.118 & 0.128 & 0.134 & 0.138 & 0.126 & 0.149 & 0.172 & 0.237 \\
          & & MAE &  0.230 & 0.245 & 0.254 & 0.284 & 0.226 & 0.232 & 0.235 & 0.240 & 0.232 & 0.247 & 0.265 & 0.316 \\
          & \multirow{2}{*}{\textbf{+\framework}} & MSE & \textbf{0.109} & \textbf{0.120} & \textbf{0.144} & \textbf{0.194} & \textbf{0.107} & \textbf{0.124} & \textbf{0.124} & \textbf{0.136} & \textbf{0.099} & \textbf{0.123} & \textbf{0.147} & \textbf{0.196} \\
          & & MAE & \textbf{0.216} & \textbf{0.228} & \textbf{0.247} & \textbf{0.282} & \textbf{0.215} & \textbf{0.230} & \textbf{0.231} & \textbf{0.245} & \textbf{0.204} & \textbf{0.224} & \textbf{0.242} & \textbf{0.278} \\
         \hline
          \multirow{6}{*}{\rotatebox{90}{Stationary}} & \multirow{2}{*}{Original} & MSE & 0.120 & 0.143 & 0.156 & 0.220 & 0.109 & 0.116 & 0.129 & 0.139 & 0.151 & 0.180 & 0.252 & 0.223 \\
          & & MAE  & 0.222 & 0.242 & 0.252 & 0.300 & 0.214 & 0.220 & 0.230 & 0.240 & 0.235 & 0.247 & 0.262 & 0.285 \\
          & \multirow{2}{*}{\textbf{+\ms}} & MSE & 0.118 & 0.143 & 0.156 & 0.208 & 0.104 & 0.115 & 0.123 & 0.136 & 0.134 & 0.160 & 0.191 & 0.231 \\
          & & MAE &  0.219 & 0.241 & 0.252 & 0.285 & 0.209 & 0.218 & 0.223 & 0.234 & 0.224 & 0.237 & 0.251 & 0.289 \\
          & \multirow{2}{*}{\textbf{+\framework}} & MSE & \textbf{0.101} & \textbf{0.131} & \textbf{0.153} & \textbf{0.190} & \textbf{0.096} & \textbf{0.114} & \textbf{0.125} & \textbf{0.135} & \textbf{0.103} & \textbf{0.144} & \textbf{0.184} & \textbf{0.217} \\
          & & MAE  & \textbf{0.206} & \textbf{0.229} & \textbf{0.245} & \textbf{0.273} & \textbf{0.199} & \textbf{0.217} & \textbf{0.229} & \textbf{0.243} & \textbf{0.200} & \textbf{0.220} & \textbf{0.245} & \textbf{0.278} \\
         \hline
          \multirow{6}{*}{\rotatebox{90}{Crossformer}} & \multirow{2}{*}{Original} & MSE & 0.159 & 0.233 & 0.275 & 0.315 & 0.149 & 0.216 & 0.230 & 0.276 & 0.141 & 0.162 & 0.199 & 0.261 \\ 
        & & MAE & 0.270 & 0.319 & 0.351 & 0.383 & 0.261 & 0.320 & 0.324 & 0.369 & 0.253 & 0.269 & 0.306 & 0.355 \\ 
        & \multirow{2}{*}{\textbf{+\ms}} & MSE & 0.134 & 0.179 & 0.217 & 0.264 & 0.133 & 0.171 & 0.186 & 0.240 & 0.112 & 0.136 & 0.156 & 0.196 \\ 
        & & MAE &   0.237 & 0.270 & 0.298 & 0.335 & 0.237 & 0.270 & 0.284 & 0.329 & 0.220 & 0.235 & 0.252 & 0.289 \\
        & \multirow{2}{*}{\textbf{+\framework}} & MSE & \textbf{0.119} & \textbf{0.166} & \textbf{0.189} & \textbf{0.223} & \textbf{0.114} & \textbf{0.139} & \textbf{0.161} & \textbf{0.171} & \textbf{0.088} & \textbf{0.108} & \textbf{0.134} & \textbf{0.171}\\
        & & MAE  & \textbf{0.217} & \textbf{0.250} & \textbf{0.265} & \textbf{0.293} & \textbf{0.215} & \textbf{0.236} & \textbf{0.258} & \textbf{0.275} & \textbf{0.190} & \textbf{0.206} & \textbf{0.222} & \textbf{0.251} \\
          \hline
    \end{tabular}
    \label{tab:pems_ablation_results}
\end{table}



\end{document}